\documentclass[10pt,journal,cspaper,compsoc]{IEEEtran}

\ifCLASSOPTIONcompsoc
\else
  \usepackage{cite}
\fi

\ifCLASSINFOpdf
\else
  \usepackage[dvips]{graphicx}
\fi
\usepackage{epsfig}

\usepackage[cmex10]{amsmath}
\usepackage{algorithmic}

\usepackage{array}

\usepackage{mdwmath}
\usepackage{mdwtab}
\usepackage{eqparbox}
\usepackage{multirow}

\usepackage{url}
\usepackage{amssymb}
\usepackage{amsthm,url,balance}
\usepackage{color}
\usepackage[T1]{fontenc}

\usepackage{algorithm}
\usepackage{algorithmic}
\usepackage[ruled,vlined,algo2e]{algorithm2e}
\providecommand{\SetAlgoLined}{\SetLine}

\usepackage{times}
\usepackage{epsfig}
\usepackage{graphicx}

\usepackage{algorithm}
\usepackage{algorithmic}
\usepackage{amsfonts}
\usepackage{amsthm}  
\usepackage{wasysym}
\usepackage{color}
\usepackage{dsfont}	
\usepackage{wrapfig}

\usepackage{footnote}

\theoremstyle{definition}

\DeclareMathOperator*{\argmax}{arg\,max}

\newcommand{\resp}{{\it resp. }}
\newcommand{\reid}{{re-id }}
\newcommand{\Rid}{{Re-id }}
\newcommand{\ie}{{\it i.e. }}
\newcommand{\etal}{{\it et al. }}
\newcommand{\eg}{{\it e.g. }}
\newcommand{\etc}{{\it etc. }}

\newcommand{\rmnum}[1]{\romannumeral #1}

\begin{document}
\title{PRISM: Person Re-Identification via Structured Matching}

\author{Ziming~Zhang,
        and Venkatesh Saligrama,~\IEEEmembership{Member,~IEEE}% <-this % stops a space
\IEEEcompsocitemizethanks{\IEEEcompsocthanksitem Dr. Z. Zhang and Prof. V. Saligrama are currently with the Department of Electrical and Computer Engineering, Boston University, Boston, MA 02215, US.\protect\\
E-mail: zzhang14@bu.edu, srv@bu.edu}% <-this % stops a space
\thanks{}}

% note the % following the last {\it i.e.}EEmembership and also \thanks -
% Spaces after {\it i.e.}EEmembership other than the last one are OK (and needed) as

\markboth{IEEE Transaction on Pattern Analysis and Machine Intelligence}%
{Zhang & Saligrama: PRISM: Person Re-Identification via Structured Matching}

%{\it i.e.}EEpubid{0000--0000/00\$00.00~\copyright~2007 IEEE}
%{\it i.e.}EEpubid{\makebox[\columnwidth]{\hfill 0000--0000/00/\$00.00~\copyright~2007 IEEE}%
% Remember, if you use this you must call {\it i.e.}EEpubidadjcol in the second

% PRIOR to the title within the {\it i.e.}EEcompsoctitleabstractindextext IEEEtran
\IEEEcompsoctitleabstractindextext{%
\begin{abstract}
Person re-identification (re-id), an emerging problem in visual surveillance, deals with maintaining entities of individuals whilst they traverse various locations surveilled by a camera network. From a visual perspective re-id is challenging due to significant changes in visual appearance of individuals in cameras with different pose, illumination and calibration. Globally the challenge arises from the need to maintain structurally consistent matches among all the individual entities across different camera views. We propose PRISM, a structured matching method to jointly account for these challenges. We view the global problem as a weighted graph matching problem and estimate edge weights by learning to predict them based on the co-occurrences of visual patterns in the training examples. These co-occurrence based scores in turn account for appearance changes by inferring likely and unlikely visual co-occurrences appearing in training instances. We implement PRISM on single shot and multi-shot scenarios. PRISM uniformly outperforms state-of-the-art in terms of matching rate while being computationally efficient.  

\end{abstract}
% not want either math or citations to appear in the abstract.

% Note that keywords are not normally used for peer review papers.
}

\maketitle

% abstract/keywords data, the IEEEcompsoctitleabstractindextext text will
% here as IEEEdisplaynotcompsoctitleabstractindextext when compsoc mode
% is not selected <OR> if conference mode is selected - because compsoc
% conference papers position the abstract like regular (non-compsoc)
\IEEEdisplaynotcompsoctitleabstractindextext

% \fi
\IEEEpeerreviewmaketitle

%%%%%%%%% BODY TEXT
\section{Introduction}\label{sec:intr}
\IEEEPARstart{M}any surveillance systems require autonomous long-term behavior monitoring of pedestrians within a large camera network. 
%can effectively reduce security-related human efforts and thus has drawn a lot of attention. 
One of the key issues in this task is {\em person re-identification} ({\it re-id}), which deals with as to how to maintain entities of individuals as they traverse through diverse locations that are surveilled by different cameras with non-overlapping camera views. As in the literature, in this paper we focus on finding entity matches between two cameras.

\Rid presents several challenges. From a vision perspective, camera views are non-overlapping and so conventional tracking methods are not helpful. Variation in appearance between the two camera views is so significant --- due to the arbitrary change in view angles, poses, illumination and calibration --- that features seen in one camera are often missing in the other. Low resolution of images for \reid makes biometrics based approaches often unreliable~\cite{Vezzani:2013:PRS:2543581.2543596}. Globally, the issue is that only a subset of individuals identified in one camera (location) may appear in the other.

%More generally, \emph{open-world} \reid~\cite{cancela2014,gong2014re,DBLP:dblp_conf/cvpr/ZhengGX12} presents further challenges including loss of tracks within one camera or missing matches on account of new entering entities. 
%
%Finally, face recognition based techniques~~\cite{Vezzani:2013:PRS:2543581.2543596} cannot be applied on account of the the relatively poor camera resolution in many surveillance systems.
%
\begin{figure}[t]
\begin{minipage}[b]{0.41\linewidth}
 \begin{center}
 \centerline{\includegraphics[width=\columnwidth]{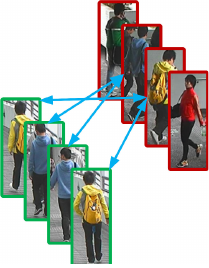}}
 \centerline{\footnotesize{(a)}}
 \end{center}
% \vspace*{-5mm}
\end{minipage}
\begin{minipage}[b]{0.58\linewidth}
\begin{center}
\centerline{\includegraphics[width=0.6\columnwidth]{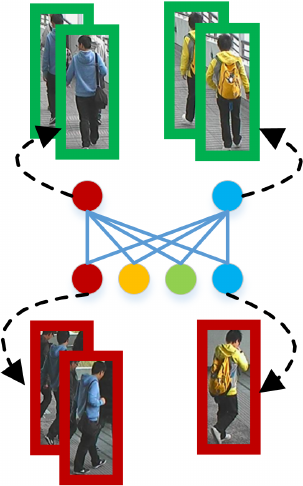}}
 \centerline{\footnotesize{(b)}}
\end{center} 
\end{minipage}
%\vspace{-5mm}
\caption{\footnotesize{(a) Illustration of re-id, where color red and green label the images from two different camera views, and arrows indicate entity matches. (b) Illustration of the weighted bipartite graph matching problem for (a), where each row denotes a camera view, each node denotes a person entity, different colors denote different entities, and the edges are weighted by 0 or 1, indicating missing matches or same entities. Each entity per view can be associated with single or multiple images.}}\label{fig:sm-intuition}
%\vspace{-5mm}
\end{figure}

\begin{figure*}
\begin{center}
\centerline{\includegraphics[width=0.95\linewidth]{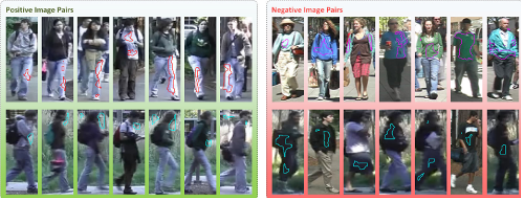}}
\caption{\footnotesize{Illustration of visual word co-occurrence in positive image pairs (\ie two images from different camera views per column belong to a {\em same} person) and negative image pairs (\ie two images from different camera views per column belong to {\em different} persons). For positive (or negative) pairs, in each row the enclosed regions are assigned the same visual word.}}\label{fig:intuition}
\end{center}
%\vspace{-1mm}
\end{figure*}

We propose, PRISM, a structured matching method for re-id. PRISM is a {\em weighted bipartite matching} method that simultaneously identifies potential matches between individuals viewed in two different cameras. % PRISM is naturally well-suited for many open-world \reid surveillance contexts \cite{cancela2014,gong2014re,DBLP:dblp_conf/cvpr/ZhengGX12} such as airports, where multiple entities are viewed at any given time. 
Fig. \ref{fig:sm-intuition}(a) illustrates \reid with two camera views, where 4 images labeled by green form the so-called probe set, and 4 entities labeled by red form the so-called gallery set. Graph matching requires edge weights, which correspond to similarity between entities viewed from two different cameras.

We learn to estimate edge weights from training instances of manually labeled image pairs. We formulate the problem as an instance of structured learning \cite{taskar2005learning} problem. While structured learning has been employed for matching text documents, \reid poses new challenges. Edge weights are obtained as a weighted linear combination of basis functions. For texts these basis functions encode shared or related words or patterns (which are assumed to be known a priori) between text documents. The weights for the basis functions are learned from training data. In this way during testing edge weights are scored based on a weighted combination of related words. In contrast, visual words (\ie vector representations of appearance information, similar to the words in texts) suffer from well known visual ambiguity and spatial distortion. This issue is further compounded in the \reid problem where visual words exhibit significant variations in appearance due to changes in pose, illumination, \etc 

%
%which can efficiently capture the pairwise co-occurrence statistics of visual words between different images. 

To handle the visual ambiguity and spatial distortion, we propose new basis functions based on co-occurrence of different visual words. We then estimate weights for different co-occurrences from their statistics in training data. While co-occurrence based statistics has been used in some other works~\cite{DBLP:conf/avss/BanerjeeN11,Galleguillos2008,Ladicky:2010:GCB:1888150.1888170}, ours has a different purpose. We are largely motivated by the observation that the co-occurrence patterns of visual codewords behave similarly for images from different views. In other words, the transformation of target appearances can be statistically inferred through these co-occurrence patterns. As seen in Fig. \ref{fig:intuition}, we observe that some regions are distributed similarly in images from different views and robustly in the presence of large cross-view variations. These regions provide important discriminant co-occurrence patterns for matching image pairs. For instance, statistically speaking, the first column of positive image pairs shows that ``white'' color in Camera 1 can change to ``light blue'' in Camera 2. However, ``light blue'' in Camera 1 can hardly change to ``black'' in Camera 2, as shown in the first column of negative image pairs.

In our previous work \cite{ZhangECCV2014}, we proposed a novel visual word co-occurrence model to capture such important patterns between images. We first encode images with a sufficiently large codebook to account for different visual patterns. Pixels are then matched into codewords or visual words, and the resulting spatial distribution for each codeword is embedded to a kernel space through {\it kernel mean embedding}~\cite{DBLP:conf/alt/SmolaGSS07} with latent-variable conditional densities \cite{Jebara:2004:PPK:1005332.1016786} as kernels. The fact that we incorporate the spatial distribution of codewords into appearance models provides us with locality sensitive co-occurrence measures. Our approach can be also interpreted as a means to {\em transfer} the information (\eg pose, illumination, and appearance) in image pairs to a common latent space for meaningful comparison. 

In this perspective appearance change corresponds to transformation of a visual word viewed in one camera into another visual word in another camera. Particularly, our method does not assume any smooth appearance transformation across different cameras. Instead, our method learns the visual word co-occurrence pattens statistically in different camera views to predict the identities of persons. The structured learning problem in our method is to determine important co-occurrences while being robust to noisy co-occurrences.

In summary, our main contributions of this paper are:
\begin{itemize}
\item We propose a new structured matching method to simultaneously identify matches between two cameras that can deal with both single-shot and multi-shot scenarios in a unified framework;
\item We account for significant change in appearance design of new basis functions, which are based on visual word co-occurrences \cite{ZhangECCV2014};
%\item \vspace{-2mm} 
%(3) Our method can deal with single-shot, multi-shot, and open-world scenarios in one framework.\\
\item We outperform the state-of-the-art significantly on several benchmark datasets, with good computational efficiency in testing. 
\end{itemize}
\subsection{Related Work}\label{ssec:rw}
While \reid has received significant interest~\cite{Vezzani:2013:PRS:2543581.2543596,doretto2011appearance,wang2014person}, much of this effort can be viewed as methods that seek to {\it classify} each probe image into one of gallery images. Broadly \reid literature can be categorized into two themes with one focusing on cleverly designing local features~\cite{ZhangECCV2014,Bak_AVSS11_MRCG,Bauml_AVSS11_evaluation,Farenzena_CVPR10_SDALF,Gheissari_CVPR06_spatiotemporal,Gray_ECCV08_ELF,conf/eccv/LiuGLL12, Ma_BMVC12_Bicov,Nguyen_NIP13_DPM,Pedagadi_CVPR13_LFDA, Prosser_BMVC10_SVR,Zhao_ICCV13_salience,zhao2013unsupervised,yang_eccv14,zheng2015query,wuviewpoint} and the other focusing on metric learning \cite{Dikmen:2010:PRL:1966111.1966152,xiong2014person,Javed:2008:MIS:1330770.1330930,Li:2012:HRT:2481913.2481917,Liu2014,Mignon_CVPR12_PCCA,porikli2003inter,Wei-ShiZheng:2011:PRP:2191740.2192190,Zheng_PAMI13_RDC,chen2015similarity,liao2014joint,Xu_ICCV13_template,Liu_ICCV13_POP}. Typically local feature design aims to find a re-id specific representation based on the some properties among the data in re-id, \eg symmetry and centralization of pedestrians in images \cite{Farenzena_CVPR10_SDALF}, color correspondences in images from different cameras \cite{zheng2015query,yang_eccv14}, spatial-temporal information in re-id videos/sequences \cite{Bauml_AVSS11_evaluation,Gheissari_CVPR06_spatiotemporal}, discriminative image representation \cite{ZhangECCV2014,Bak_AVSS11_MRCG,Ma_BMVC12_Bicov}, viewpoint invariance prior \cite{wuviewpoint}. Unlike these approaches that attempt to match local features our method attempts to learn changes in appearance or features to account for visual ambiguity and spatial distortion. On the other hand, metric learning aims to learn a better similarity measure using, for instance, transfer learning \cite{Li:2012:HRT:2481913.2481917}, dictionary learning \cite{Liu2014}, distance learning/comparison \cite{Mignon_CVPR12_PCCA,Wei-ShiZheng:2011:PRP:2191740.2192190,Zheng_PAMI13_RDC}, similarity learning \cite{chen2015similarity}, dimension reduction \cite{liao2014joint}, template matching \cite{Xu_ICCV13_template}, active learning \cite{Liu_ICCV13_POP}. In contrast to metric learning approaches that attempt to find a metric such that features from positively associated pairs are close in distance our learning algorithm learns similarity functions for imputing similarity between features that naturally undergo appearance changes.

Re-ID can also be organized based on so called single-shot or multi-shot scenarios. For {\em single-shot learning}, each entity is associated with only one single image, and \reid is performed based on every single image pair. In the literature, most of the methods are proposed under this scenario. For instance, Zhao \etal \cite{zhao2014learning} proposed learning good discriminative mid-level filters for describing images. Yang \etal \cite{yang_eccv14} proposed a saliency color based image descriptor and employed metric learning with these descriptors for re-id. 
%Zhang \etal \cite{zhang_eccv14} proposed a visual word co-occurrence model.
% to deal with visual ambiguity and spatial distortion simultaneously.
For {\em multi-shot learning}, each entity is associated with at least one image, and \reid is performed based on multiple image pairs. How to utilize the redundant information in multiple images is the key difference from single-shot learning. Wu \etal \cite{LCRNP_FCV14} proposed a locality-constrained collaboratively regularized nearest point model to select images for generating decision boundaries between different entities, which are represented as sets of points in the feature space. Bazzani \etal \cite{Bazzani:2012:MPR:2161002.2161235} propose a new image representation by focusing on the overall chromatic content and the presence of recurrent local patches.

%For {\em cross-domain learning}, a model is learned from one {\em known} dataset/camera-pair, and tested on another {\em unknown} dataset/camera-pair without change. Typically it is assumed that two datasets can be aligned in certain common space so that the information in one can be transfered to another. Li and Wang \cite{conf/cvpr/LiW13} proposed aligning the visual features of image pairs from different views locally in a common feature space and then matching them with softly assigned metrics. Ma \etal \cite{jinhuama} proposed a Domain Transfer Ranked Support Vector Machines (DTRSVM) method by assuming the difference between the means of matched and unmatched image features in the source domain is similar to that in the target domain. 

Our work in contrast deals with these different scenarios within one framework. In addition we allow for no matches for some entities and can handle cases where the numbers of entities in both probe and gallery sets are different. Meanwhile, our basis function can handle both single-shot and multi-shot learning directly while accounting for appearance changes. 

While special cases of our method bears similarity to Locally-adaptive Decision Functions (LADF) described in \cite{conf/cvpr/LiCLHCS13}, they are fundamentally different. LADF proposes a second-order (\ie quadratic) decision function based on metric learning. In contrast we compute similarities between entities and do not need to impose positive semidefinite conditions during training. Our method is also related to \cite{Das2014} where an integer optimization method was proposed to enforce network consistency in \reid during testing, \ie maintaining consistency in \reid results across the network. For instance, a person $A$ from camera view 1 matches a person $B$ from view 2, and $A$ also matches a person $C$ from view 3, then based on consistency $B$ should match $C$ as well. This network consistency helps improve the camera pairwise \reid performance between all the individual camera pairs. In contrast, graph-structure is integral to both training and testing in our proposed approach. We \emph{learn-to-estimate} bipartite graph structures during testing by pruning the feasible solution space based on our a priori knowledge on correct matching structures. Recently, Paisitkriangkrai \etal \cite{2015arXiv150301543P} and Liu \etal \cite{LiuWACV2015} proposed utilizing structured learning to integrate the metric/color model ensembles, where structured learning is taken as a means to enhance the \reid performance of each individual model. In contrast, we consider structured learning as a way to learn the classifier, working with our own features for re-id.

To summarize our contributions, our method learns to assign weights to pairs of instances using globally known feasible assignments in training data. Unlike text data or other conventional approaches our weights incorporate appearance changes and spatial distortion. We express the weights as a linear combination of basis functions, which are the set of all feasible appearance changes (co-occurrences). Our decision function is a weighting function that weights different co-occurrences. During training, our structural constraints induce higher scores on ground-truth assignments over other feasible assignments. During testing, we enforce a globally feasible assignment based on our learned co-occurrence weights.

Very recently, open-world \reid \cite{gong2014re,DBLP:dblp_conf/cvpr/ZhengGX12,cancela2014} has been introduced, where persons in each camera may be only partially overlapping and the number of cameras, spatial size of the environment, and number of people may be unknown and at a significantly larger scale. Recall that the goal of this paper is to identify the persons given aligned images, which are the cases in most person re-identification benchmark datasets, while open-world re-id this is more a system level concept that must deal with issues such as person detection, tracking, re-id, data association, \etc Therefore, open-world re-id is out of scope of our current work.

Structured learning has been also used in the object tracking literature (\eg \cite{hare2011struck}) for data association. The biggest difference, however, between our method and these tracking methods is that in our \reid cases, we do not have any temporal or location information with data, in general, which leads to totally different goals: our method aims to find the correct matches among the entities using structured matching in testing based on only the appearance information, while in tracking the algorithms aim to associate the same object with small appearance variations in two adjacent frames locally. 

The rest of this paper is organized as follows: Section \ref{sec:method} explains our structured prediction method in detail. Section \ref{sec:implementation} lists some of our implementation details. Section \ref{sec:exp} reports our experimental results on the benchmark datasets. We conclude the paper in Section \ref{sec:con}.
\section{PRISM}\label{sec:method}
In this paper we focus on two camera \reid problems, as is common in the literature. In the sequel we present an overview of our proposed method.% we give an overview of PRISM and then present the various components of training and testing in subsequent sections. %In Section \ref{ssec:sm}, we explain how to perform entity-level structured matching. In Section \ref{ssec:patch_matching}, we define our basis function for entity similarity measure. In Section \ref{ssec:sl},  we present the details of the structured learning formulation for PRISM.
%In Section \ref{ssec:feature_transformation}, we introduce a second bipartite graph matching formulation for unsupervised learning of linear feature transformation matrices across different datasets.  
%

\subsection{Overview}\label{ssec:pre}

\begin{figure}[t]
%\begin{minipage}[b]{0.25\linewidth}
% \begin{center}
 \centerline{\includegraphics[width=\columnwidth]{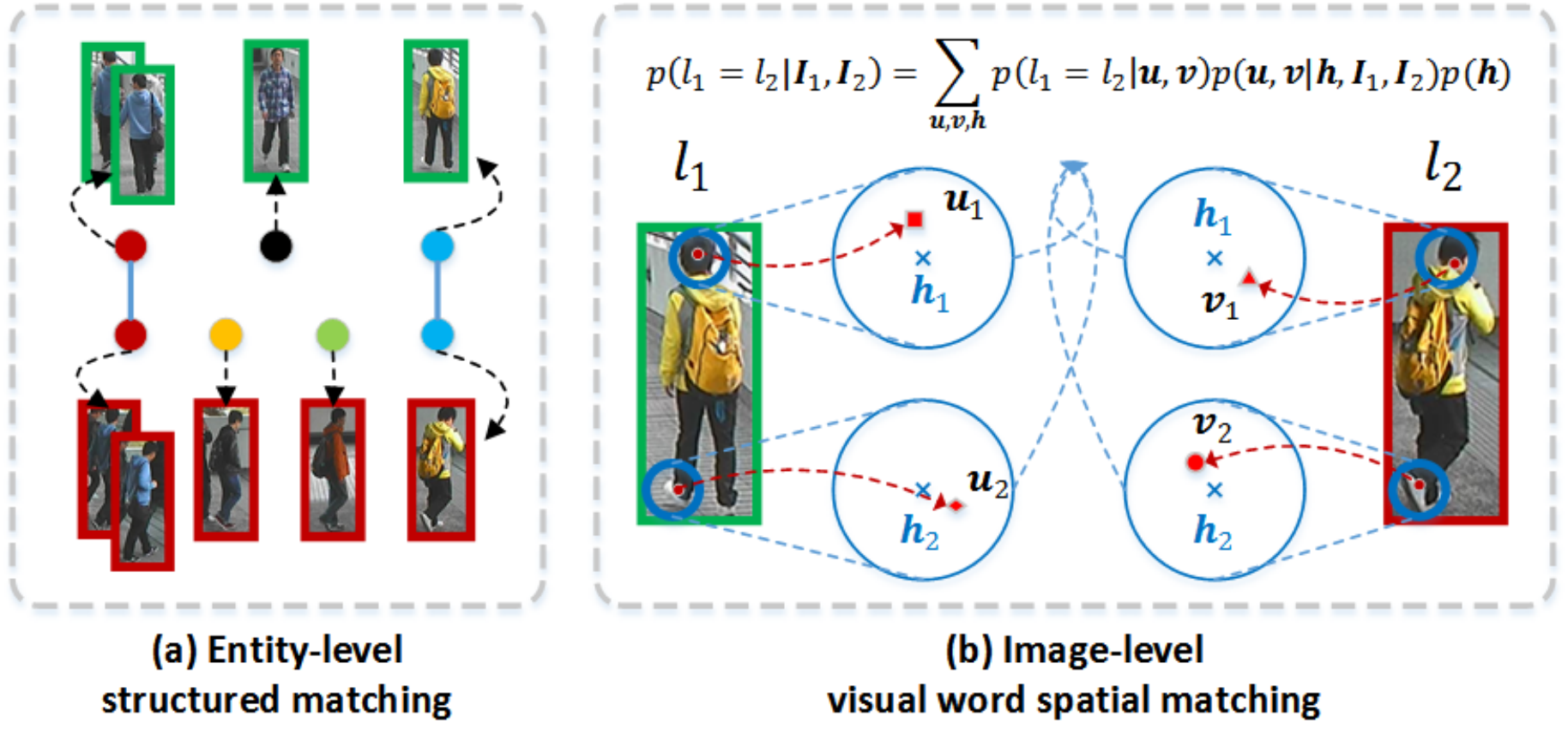}}
% \centerline{\footnotesize{(a) entity matching}}
% \end{center}
% \vspace*{-5mm}
%\end{minipage}
%\begin{minipage}[b]{0.74\linewidth}
%\begin{center}
%\centerline{\includegraphics[width=0.6\columnwidth]{basis-function.pdf}}
% \centerline{\footnotesize{(b) Image matching}}
%\end{center} 
%\end{minipage}
%\vspace{-5mm}
\caption{\footnotesize{Overview of our method, PRISM, consisting of two levels where (a) entity-level structured matching is imposed on top of (b) image-level visual word deformable matching. In (a), each color represents an entity, and this example illustrates the general situation for re-id, including single-shot, multi-shot, and no match scenarios. In (b), the idea of visual word co-occurrence for measuring image similarities is illustrated in a probabilistic way, where $l_1,l_2$ denote the person entities, $\mathbf{u}_1,\mathbf{u}_2,\mathbf{v}_1,\mathbf{v}_2$ denote different visual words, and $\mathbf{h}_1,\mathbf{h}_2$ denote two locations.}}\label{fig:overview}
%\vspace{-3mm}
\end{figure}

Let us first describe the problem we often encounter during testing. We are given $N_1$ probe entities (Camera 1) that are to be matched to $N_2$ gallery entities (Camera 2). Fig.~\ref{fig:overview} depicts a scenario where entities may be associated with a single image (single-shot), multiple images (multi-shot) and be unmatched to any other entity in the probe/gallery (\eg ``black'', ``orange'', and ``green'' entities in (a)). {\em Existing methods could fail here for the reason that entities are matched independently based on pairwise similarities between the probes and galleries leading to the possibility of matching multiple probes to the same entity in the gallery.} Structured matching is a framework that can address some of these issues. 

To build intuition, consider $\bar y_{ij}$ as a binary variable denoting whether or not there is a match between $i^{th}$ probe entity and $j^{th}$ gallery entity, and $s_{ij}$ as their similarity score. Our goal is to predict the structure, $\bar{\mathbf{y}}$, by seeking a maximum bipartite matching:
\begin{align} \label{e.struct}%\noname
\max_{\forall i,\forall j, \bar{y}_{ij}\in [0,1]} \; \sum_{i,j}\bar{y}_{ij}s_{ij},\,\,
\mbox{s.t.} \;\; \mathbf{\bar{y}}=[\bar{y}_{ij}]_{\forall i,\forall j}\in {\cal Y}
%\sum_j\bar{y}_{ij}\leq r_i, %\nonumber
%\mbox{s.t.} & \; \forall j, \; 0\leq \bar{y}_{ij}\leq 1, \; %\sum_j\bar{y}_{ij}=1, \nonumber
\end{align}
where ${\cal Y}$ could be the sub-collection of bipartite graphs accounting for different types of constraints. For instance, ${\cal Y}=\{\mathbf{\bar{y}}\mid \forall i, \sum_j\bar{y}_{ij}\leq r_i,\,\forall j, \sum_i\bar{y}_{ij}\leq g_j\}$ would account for the relaxed constraint to identify {\em at most} $r_i$ potential matches from the gallery set for probe $i$, and {\em at most} $g_j$ potential matches from the probe set for gallery $j$. Hopefully the correct matches are among them. \\%Note that our setting allows for more general graphs, as we will see later, and can model some open-world scenarios by imposing other constraints.% then the degree of node $i$ in the graph is $r_i$.
%where $\bar{\mathbf{y}}_i=\{\bar{y}_{ij}\}$ denotes the predicted matching weights for entity $i$, and $\forall j$ denotes the index of an entity in the gallery.

\noindent
{\bf Learning Similarity Functions:} Eq.~\ref{e.struct} needs similarity score $s_{ij}$ for every pair of probe $i$ and gallery $j$, which is a priori unknown and could be arbitrary. Therefore, we seek similarity models that can be learned from training data based on minimizing some {\em loss function}. %This immediately raises the question of as to how to evaluate different similarity models?

Structured learning~\cite{taskar2005learning} formalizes loss functions for learning similarity models that are consistent with testing goals as in Eq.~\ref{e.struct}. To build intuition, consider the example of text documents, where each document is a collection of words chosen from a dictionary ${\cal V}$. Let $\mathcal{D}_i,\,\mathcal{D}_j$ be documents associated with probe $i$ and gallery $j$. Let $\mathcal{D}$ denote the tuple of all training probe and gallery documents. A natural similarity model is one based on \emph{shared-words} in the two documents, namely, $s_{ij} = \sum_{v \in {\cal V}} w_v \mathbf{1}_{\{v \in \mathcal{D}_i \cap v \in \mathcal{D}_j\}}$. $w_v$ denotes the importance of word $v$ in matching any two arbitrary documents. The learning problem reduces to learning the weights $w_v$ for each word from training instances that minimizes some loss function.
%
%
%We can benefit from training data by considering the ground truth as a structured matching in parallel to what we see during testing~\cite{}. 
%For text documents we encode entities from a vocabulary of codewords $v \in {\cal V}$ and associate entities with a subset of code-words $R_i, \,R_j$. 
%
%We can then model the similarity function as a weighted superposition of co-occurring words, namely, $s_{ij} = \sum_{v \in {\cal V}} w_v \mathbf{1}_{\{v \in R_i \cap \in R_j\}}$. 
%
A natural loss function is one that reflects our objectives in testing. In particular, substituting this similarity model in Eq.~\ref{e.struct}, we obtain $\sum_{i,j}\bar{y}_{ij}s_{ij} = \sum_{v \in {\cal V}} w_v \sum_{i,j}\bar{y}_{ij}\mathbf{1}_{\{v \in \mathcal{D}_i\cap v \in \mathcal{D}_j\}}$. We denote as $f_v(\mathcal{D}, \bar{\mathbf{y}})=\sum_{i,j}\bar{y}_{ij}\mathbf{1}_{\{v \in \mathcal{D}_i\cap v \in \mathcal{D}_j\}}$ the {\it basis function} associated with word $v$. It measures the frequency with which word $v$ appears in matched training instances. A loss function must try to ensure that,
\begin{equation} \label{e.lossfunc}
\sum_{v \in {\cal V}} w_vf_v(\mathcal{D}, \mathbf{y}) \geq \sum_{v \in {\cal V}} w_vf_v(\mathcal{D}, \bar{\mathbf{y}}),\,\, \forall\,\bar{\mathbf{y}}\in {\cal Y} 
\end{equation}
where $\bar{\mathbf{y}}$ is any bipartite matching and $\mathbf{y}$ is the ground-truth bipartite matching. Hinge losses can be used to penalize violations of Eq.~\ref{e.lossfunc}. Note that such loss functions only constrain the weights so that they perform better only on alternative bipartite matchings, rather than any arbitrary $\bar{\mathbf{y}}$.\\

\noindent
{\bf Similarity Models for Re-ID} are more complex relative to the example above. First, we typically have images and need a way to encode images into visual words. Second, visual words are not typically shared even among matched entities. Indeed a key challenge here is to account for significant visual ambiguity and spatial distortion, due to the large variation in appearance of people from different camera views.

We propose similarity models based on cross-view visual word co-occurrence patterns. {\it Our {\bf key insight} is that aspects of appearance that are transformed in predictable ways, due to the static camera view angles, can be statistically inferred through pairwise co-occurrence of visual words.} In this way, we allow the same visual concepts to be mapped into different visual words, and account for visual ambiguity.

We present a probabilistic approach to motivate our similarity model in Fig. \ref{fig:overview}(b). We let the similarity $s_{ij}$ be equal to the probability that two entities are identical, \ie,
\begin{align}\label{eqn:probabilistic}
& s_{ij}\stackrel{\Delta}{=}p(\bar{y}_{ij}=1|\mathbf{I}_i^{(1)},\mathbf{I}_j^{(2)}) \\
&=\sum_{\mathbf{u}\in\mathcal{U},\mathbf{v}\in\mathcal{V},\mathbf{h}\in\Pi}p(\bar{y}_{ij}=1|\mathbf{u},\mathbf{v})p(\mathbf{u},\mathbf{v}|\mathbf{h},\mathbf{I}_i^{(1)},\mathbf{I}_j^{(2)})p(\mathbf{h})\nonumber\\
&=\sum_{\mathbf{u}\in\mathcal{U},\mathbf{v}\in\mathcal{V}}p(\bar{y}_{ij}=1|\mathbf{u},\mathbf{v})\left[\sum_{\mathbf{h}\in\Pi}p(\mathbf{u},\mathbf{v}|\mathbf{h},\mathbf{I}_i^{(1)},\mathbf{I}_j^{(2)})p(\mathbf{h})\right], \nonumber
\end{align}
where $\mathbf{I}_i^{(1)},\mathbf{I}_j^{(2)}$ denote two images from camera view 1 (left) and 2 (right), respectively, $\mathbf{u}\in\mathcal{U},\mathbf{v}\in\mathcal{V}$ denote the visual words for view 1 and view 2, and $\mathbf{h}\in\Pi$ denotes the shared spatial locations.

Following along the lines of the text-document setting we can analogously let $w_{uv}=p(\bar{y}_{ij}=1|\mathbf{u},\mathbf{v})$ denote the likelihood (or importance) of co-occurrence of the two visual words among matched documents. This term is data-independent and must be learned from training instances as before. The basis function, $f_{uv}(\cdot)$ is given by $\sum_{\mathbf{h}\in\Pi}p(\mathbf{u},\mathbf{v}|\mathbf{h},\mathbf{I}_i^{(1)},\mathbf{I}_j^{(2)})p(\mathbf{h})$ and must be {\it empirically} estimated. The basis function $f_{uv}(\cdot)$ measures the frequency with which two visual words co-occur after accounting for spatial proximity. The term $p(\mathbf{u},\mathbf{v}|\mathbf{h},\mathbf{I}_i^{(1)},\mathbf{I}_j^{(2)})$ here denotes the joint contribution of the visual words at location $\mathbf{h}$. To handle spatial distortion of visual words, we allow the visual words to be deformable, similar to deformable part model \cite{lsvm-pami}, when calculating their joint contribution. $p(\mathbf{h})$ denotes the importance of location $\mathbf{h}$ for prediction. 

In summary, our similarity model handles both visual ambiguity (through co-occurring visual words) and spatial distortion simultaneously. %similarity matrices for entity-level structured matching.
We learn parameters, $w_{uv}$, of our similarity model along the lines of Eq.~\ref{e.lossfunc} with analogous structured loss functions that penalize deviations of predicted graph structures from ground-truth annotated graph structures.
%Structured learning is a technique for predicting correct complex structures by minimizing the cost between ground-truth and predicted structures with certain constraints on the graphs. We utilize structured learning to learn the classifiers (\ie $p(l_1=l_2|\mathbf{u},\mathbf{v})$) discriminatively, and predict the graph structures $\bar{\mathbf{y}}$.
In the following sections we present more details of the different components of our proposed approach.

\subsection{Structured Matching of Entities in Testing}\label{ssec:sm}

Now let us consider the \reid problem as a bipartite graph matching problem, where all the entities are represented as the nodes in the graph, forming two sets of nodes for the probe set and the gallery set, respectively, and the matching relations are represented as the edges with weights from $\{0, 1\}$, as illustrated in Fig. \ref{fig:overview}(a). 

The {\bf key insight} of our structured matching in testing is to narrow down the feasible solution space for structured prediction in weighted bipartite graph matching based on the prior knowledge on correct matching structures.

During training, since the bipartite graph can be defined based on the training data, the degree of each node can be easily calculated. But during testing we have to predict the degree of each node. Usually the node degrees in the probe can be given beforehand. For instance, we would like to find {\em at most} $r_i$ entity matches in the gallery set for entity $i$ in the probe set so that hopefully the correct match is among them, then the degree of node $i$ in the graph is $r_i$. However, this is not the case for the nodes in the gallery. 

Therefore, without any prior on the graph structure during testing, we enforce it to have the following structural properties, which are very reasonable and practical:
\begin{itemize}
\item[(1)] All the entities in either gallery or probe set are {\em different} from each other, and every test entity $i$ in the probe can be {\em equally} matched with any entity in the gallery. It turns out that in this way we actually maximize the matching likelihood for the test entity $i$.
%\begin{align}
%\max_{\bar{\mathbf{y}}_i} & \; \prod_j\bar{y}_{ij}\\
%\mbox{s.t.} & \; \forall j, \; 0\leq \bar{y}_{ij}\leq 1, \; \sum_j\bar{y}_{ij}=1, \nonumber
%\end{align}
%where $\bar{\mathbf{y}}_i=\{\bar{y}_{ij}\}$ denotes the predicted matching weights for entity $i$, and $\forall j$ denotes the index of an entity in the gallery.
%\vspace{-3mm}
\item[(2)] We constrain the nodes in the gallery set to have {\em similar} degrees. This helps avoid the mismatched cases such as multiple entities in the probe being matched with the same entity in the gallery.
\end{itemize}

\begin{figure}[t]
\begin{minipage}[b]{0.41\linewidth}
 \begin{center}
 \centerline{\includegraphics[width=0.88\columnwidth]{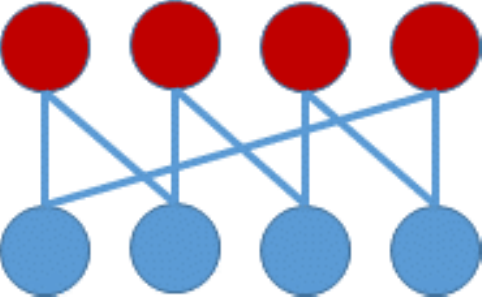}}
 \centerline{\footnotesize{(a)}}
 \end{center}
% \vspace*{-5mm}
\end{minipage}
\begin{minipage}[b]{0.58\linewidth}
\begin{center}
\centerline{\includegraphics[width=0.8\columnwidth]{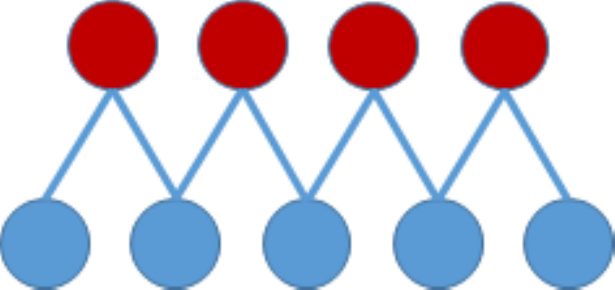}}
 \centerline{\footnotesize{(b)}}
\end{center} 
\end{minipage}
%\vspace{-10mm}
\caption{\footnotesize{Illustration of our predicted bipartite matching graphs, where red and blue nodes represent the probe and gallery sets, respectively. Both graphs in (a) and (b) are examples which satisfy the conditions (1) and (2) during testing.}}\label{fig:sm-test}
%\vspace{-5mm}
\end{figure}

We illustrate two examples for our predicted graphs during testing that satisfy the both conditions in Fig. \ref{fig:sm-test}, and we would like to find at most $r_i=2$ matches in the gallery for each probe. Then the red node degrees can be no large than 2. Accordingly, the total degree in the gallery set, where each node degree needs to be similar to others, should be the same as that in the probe set. By minimizing the entropy of the node degrees in the gallery, we can easily calculate the upper bound of the gallery node degrees in (a) as $\left\lceil\frac{4\times 2}{4}\right\rceil=2$ and in (b) as $\left\lceil\frac{4\times 2}{5}\right\rceil=2$, respectively.

%In addition to the constraint that the total node degree in the probe set is equal to that in the gallery set, 
By incorporating these node degree upper bounds, we can narrow down the feasible solution space for correct matching structures from $\{0,1\}^{N_1\times N_2}$, where $N_1$ and $N_2$ denote the numbers of nodes in the bipartite graph from view 1 (probe) and view 2 (gallery), to $\mathcal{Y}$ such that
\begin{align}\label{eqn:y}
&\mathcal{Y}=\nonumber\\
&\left\{\mathbf{y}\Big|\forall i, \forall j, y_{ij}\in \{0,1\}, \sum_jy_{ij}\leq r_i, \sum_iy_{ij}\leq\left\lceil\frac{\sum_ir_i}{N_2}\right\rceil\right\},
\end{align}
where $\forall i, r_i$ denotes the predefined degree for node $i$ in the probe, and $\lceil\cdot\rceil$ denotes the ceiling function. As we see, $r_i$ and $\left\lceil\frac{\sum_ir_i}{N_2}\right\rceil$ are used to control the node degrees in the probe and gallery, respectively, and $\left\lceil\frac{\sum_ir_i}{N_2}\right\rceil$ enforces the gallery node degrees to be similar to each other.

Then we can formulate our structured matching, \ie weighted bipartite graph matching, for re-id {\em during testing} as follows:
\begin{align}\label{eqn:test}
%\max_{\bar{\mathbf{y}}\in\mathcal{Y}} & \; \mathbf{w}^Tf(\hat{\mathcal{X}}, \bar{\mathbf{y}}) = 
\mathbf{y}^*=\argmax_{\bar{\mathbf{y}}\in\mathcal{Y}} \mathbf{w}^Tf(\mathcal{X},\bar{\mathbf{y}})=\argmax_{\bar{\mathbf{y}}\in\mathcal{Y}}\left\{\sum_{i,j}\bar{y}_{ij}\mathbf{w}^T\phi(\mathbf{x}_{ij})\right\},
\end{align}
where $\mathbf{x}_{ij}\in\mathcal{X}$ denotes an entity pair between entity $i$ in the probe and entity $j$ in the gallery, $\phi(\cdot)$ denotes the {\em similarity measure function}, $\mathbf{w}$ denotes the weight vector for measuring entity similarities, $(\cdot)^T$ denotes the matrix transpose operator, $\bar{\mathbf{y}}\in\mathcal{Y}$ denotes a matching structure from the structure set $\mathcal{Y}$, and $\mathbf{y}^*$ denotes the predicted matching structure for re-id. Note that $f(\mathcal{X},\bar{\mathbf{y}})=\sum_{i,j}\bar{y}_{ij}\phi(\mathbf{x}_{ij})$ is our {\em basis function} for re-id. 

Functionally, $\mathbf{w}$ and $\phi(\mathbf{x}_{ij})$ in Eq. \ref{eqn:test} stand for $p(\bar{y}_{ij}=1|\mathbf{u},\mathbf{v})$ and $\sum_{\mathbf{h}\in\Pi}p(\mathbf{u},\mathbf{v}|\mathbf{h},\mathbf{I}_i^{(1)},\mathbf{I}_j^{(2)})p(\mathbf{h})$ in Eq. \ref{eqn:probabilistic}, respectively. $\forall i,\forall j, \mathbf{w}^T\phi(\mathbf{x}_{ij})$ defines the edge weight between node $i$ and node $j$ in the bipartite graph. Our method learns the 0/1 assignments for the edges under the structural conditions, so that the total weight over the bipartite graph is maximized. Given these edge weights, we can utilize linear programming to solve Eq. \ref{eqn:test}, and then threshold the solution to return the 0/1 assignments. Notice that our structured matching can handle the general entity matching problem as illustrated in Fig. \ref{fig:overview}(a), which is different from conventional \reid methods.

\subsection{Similarity Models}\label{ssec:patch_matching}

Now we come to the question of as to how we define our similarity measure function $\phi(\cdot)$ in Eq. \ref{eqn:test}. Recall that our method has to deal with (1) single-shot learning, (2) multi-shot learning, (3) visual ambiguity, and (4) spatial distortion. Following Fig. \ref{fig:overview}(b), we define $\phi(\cdot)$ based on the cross-view visual word co-occurrence patterns.

\begin{figure*}[t]
\begin{center}
\centerline{\includegraphics[width=0.9\linewidth]{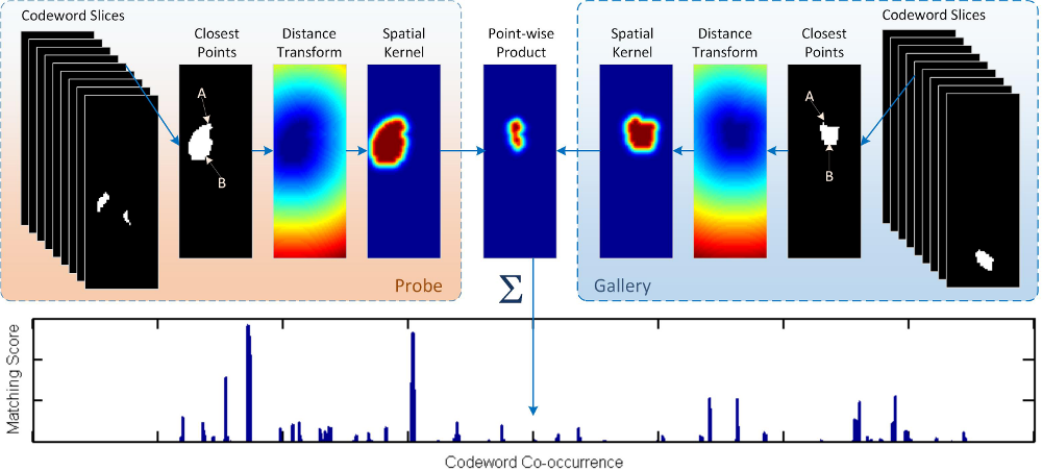}}
\caption{\footnotesize{Illustration of visual word co-occurrence model generation process in \cite{ZhangECCV2014}. Here, the white regions in the codeword slices indicate the pixel locations with the same codeword. ``A'' and ``B'' denote two arbitrary pixel locations in the image domain. And ``$\Sigma$'' denotes a sum operation which sums up all the values in the point-wise product matrix into a single value $[\phi(\mathbf{x}_{ij})]_{uv}$ in the model.}}\label{fig:matching}
\end{center}
\end{figure*}

\subsubsection{Locally Sensitive Co-occurrence \cite{ZhangECCV2014}}
We need co-occurrence models that not only account for the locality of appearance changes but also the random spatial and visual ambiguity inherent in vision problems. Recall that we have two codebooks $\mathcal{U}=\{\mathbf{u}\}$ and $\mathcal{V}=\{\mathbf{v}\}$ for view 1 and view 2, respectively. Our codebook construction is global and thus only carries information about distinctive visual patterns. Nevertheless, for a sufficiently large codebook distinctive visual patterns are mapped to different elements of the codebook, which has the effect of preserving local visual patterns. Specifically, we map each pixel at 2D location $\boldsymbol{\pi} \in \Pi$ of image $\mathbf{I}$ in a view into one codeword to cluster these pixels.

To emphasize local appearance changes, we look at the spatial distribution of each codeword. Concretely, we let $\Pi_u=\mathcal{C}(\mathbf{I},\mathbf{u})\subseteq\Pi$ ({\it resp.} $\Pi_v=\mathcal{C}(\mathbf{I},\mathbf{v})\subseteq\Pi$) denote the set of pixel locations associated with codeword $\mathbf{u}$ ({\it resp.} $\mathbf{v}$) in image $\mathbf{I}$ and associate a spatial probability distribution, $p(\boldsymbol{\pi}|\mathbf{u},\mathbf{I})$ ({\it resp.} $p(\boldsymbol{\pi}|\mathbf{v},\mathbf{I})$), over this observed collection. In this way visual words are embedded into a family of spatial distributions. Intuitively it should now be clear that we can use the similarity (or distance) of two corresponding spatial distributions to quantify the pairwise relationship between two visual words. This makes sense because our visual words are spatially locally distributed and small distance between spatial distributions implies spatial locality. Together this leads to a model that accounts for local appearance changes. 

While we can quantify the similarity between two distributions in a number of ways, the kernel mean embedding \cite{DBLP:conf/alt/SmolaGSS07} method is particularly convenient for our task. The basic idea to map the distribution, $p$, into a reproducing kernel Hilbert space (RKHS), ${\cal H}$, namely, $p \rightarrow  \mu_p(\cdot) = \sum K(\cdot, \boldsymbol{\pi}) p(\boldsymbol{\pi}) \stackrel{\Delta}{=} E_p(K(\cdot, \boldsymbol{\pi}))$. For universal kernels, such as RBF kernels, this mapping is injective, {\it i.e.}, the mapping preserves the information about the distribution \cite{DBLP:conf/alt/SmolaGSS07}. In addition we can exploit the reproducing property to express inner products in terms of expected values, namely, $\langle \mu_p, \Phi \rangle = E_p(\Phi),\,\forall\, \Phi \in {\cal H}$ and obtain simple expressions for similarity between two distributions (and hence two visual words) because $\mu_p(\cdot) \in {\cal H}$.  

To this end, consider the codeword $\mathbf{u}\in\mathcal{U}$ in image $\mathbf{I}_i^{(1)}$ and the codeword $\mathbf{v}\in\mathcal{V}$ in image $\mathbf{I}_j^{(2)}$. The co-occurrence matrix (and hence the appearance model) is the inner product of visual words in the RKHS space, namely,
\begin{align}\label{eqn:kme}
&\Big[\phi(\mathbf{x}_{ij})\Big]_{uv}=\left\langle \mu_{p(\cdot \mid \mathbf{u},\mathbf{I}_i^{(1)})}, \mu_{p(\cdot \mid \mathbf{v},\mathbf{I}_j^{(2)})}\right\rangle \\
&= \sum_{\boldsymbol{\pi}_u\in\Pi}\sum_{\boldsymbol{\pi}_v\in\Pi}K(\boldsymbol{\pi}_u,\boldsymbol{\pi}_v)p(\boldsymbol{\pi}_u|\mathbf{u},\mathbf{I}_i^{(1)})p(\boldsymbol{\pi}_v|\mathbf{v},\mathbf{I}_j^{(2)}), \nonumber
\end{align}
where we use the reproducing property in the last equality and $[\cdot]_{uv}$ denotes the entry in $\phi(\mathbf{x}_{ij})$ for the codeword pair $(\mathbf{u}, \mathbf{v})$.

Particularly, in \cite{ZhangECCV2014} we proposed a {\em latent spatial kernel}. This is a type of probability product kernel that has been previously proposed \cite{Jebara:2004:PPK:1005332.1016786} to encode generative structures into discriminative learning methods. In our context we can view the presence of a codeword $\mathbf{u}$ at location $\boldsymbol{\pi}_u$ as a noisy displacement of a true latent location $\mathbf{h} \in \Pi$. The key insight here is that the spatial activation of the two codewords $\mathbf{u}$ and $\mathbf{v}$ in the two image views $\mathbf{I}_i^{(1)}$ and $\mathbf{I}_j^{(2)}$ are conditionally independent when conditioned on the true latent location $\mathbf{h}$, namely,  the joint probability factorizes into $p\{ \boldsymbol{\pi}_u,\boldsymbol{\pi}_v|\mathbf{h},\mathbf{I}_i^{(1)},\mathbf{I}_j^{(2)}\} = p\{\boldsymbol{\pi}_u|\mathbf{h}, \mathbf{I}_i^{(1)}\}p\{\boldsymbol{\pi}_v|\mathbf{h}, \mathbf{I}_j^{(2)}\}$. We denote the noisy displacement likelihoods, $p\{\boldsymbol{\pi}_u|\mathbf{h},\mathbf{I}_i^{(1)}\} = \kappa(\boldsymbol{\pi}_u,\mathbf{h})$ and $p\{\boldsymbol{\pi}_v|\mathbf{h},\mathbf{I}_j^{(2)}\} = \kappa(\boldsymbol{\pi}_v,\mathbf{h})$ for simplicity. This leads us to $K(\boldsymbol{\pi}_u,\boldsymbol{\pi}_v)=\sum_{\mathbf{h}}\kappa(\boldsymbol{\pi}_u,\mathbf{h})\kappa(\boldsymbol{\pi}_v,\mathbf{h})p(\mathbf{h})$, where $p(\mathbf{h})$ denotes the spatial probability at $\mathbf{h}$. By plugging this new $K$ into Eq. \ref{eqn:kme}, we have
\begin{align}\label{eqn:latent}
\Big[\phi(\mathbf{x}_{ij})\Big]_{uv} = & \sum_{\boldsymbol{\pi}_u\in\Pi}\sum_{\boldsymbol{\pi}_v\in\Pi}\sum_{\mathbf{h}\in\Pi}\kappa(\boldsymbol{\pi}_u,\mathbf{h})\kappa(\boldsymbol{\pi}_v,\mathbf{h})p(\mathbf{h})\\
& \cdot p(\boldsymbol{\pi}_u|\mathbf{u},\mathbf{I}_i^{(1)})p(\boldsymbol{\pi}_v|\mathbf{v},\mathbf{I}_j^{(2)}) \nonumber\\
\leq &  \sum_{\mathbf{h}}\max_{\boldsymbol{\pi}_u}\left\{\kappa(\boldsymbol{\pi}_u,\mathbf{h})p(\boldsymbol{\pi}_u|\mathbf{u},\mathbf{I}_i^{(1)})\right\}\nonumber\\
& \cdot \max_{\boldsymbol{\pi}_v}\left\{\kappa(\boldsymbol{\pi}_v,\mathbf{h})p(\boldsymbol{\pi}_v|\mathbf{v},\mathbf{I}_j^{(2)})\right\}p(\mathbf{h}),\nonumber
\end{align}
where the inequality follows by rearranging the summations and standard upper bounding techniques. Here we use an upper bound for computational efficiency, and assume that $p(\mathbf{h})$ is a uniform distribution for simplicity without further learning. The main idea here is that by introducing the latent displacement variables, we have a handle on view-specific distortions observed in the two cameras. Using different kernel functions $\kappa$, the upper bound in Eq. \ref{eqn:latent} results in different latent spatial kernel functions.

Fig. \ref{fig:matching} illustrates the whole process of generating the latent spatial kernel based appearance model given the codeword images, each of which is represented as a collection of codeword slices. For each codeword slice, the $\max$ operation is performed at every pixel location to search for the spatially closest codeword in the slice. This procedure forms a distance transform image, which is further mapped to a spatial kernel image. It allows each peak at the presence of a codeword to be propagated smoothly and uniformly. To calculate the matching score for a codeword co-occurrence, the spatial kernel from a probe image and another from a gallery image are multiplied element-wise and then summed over all latent locations. This step guarantees that our descriptor is insensitive to the noise data in the codeword images. This value is a single entry at the bin indexing the codeword co-occurrence in our descriptor for matching the probe and gallery images. As a result, we have generated a high dimensional sparse appearance descriptor. Note that we simply the computation of this model by utilizing the indicator function for $p(\boldsymbol{\pi}_u|\mathbf{u},\mathbf{I}_i^{(1)})$ and $p(\boldsymbol{\pi}_v|\mathbf{v},\mathbf{I}_j^{(2)})$, respectively. Namely, $p(\boldsymbol{\pi}_u|\mathbf{u},\mathbf{I}_i^{(1)})=1$ ({\it resp.} $p(\boldsymbol{\pi}_v|\mathbf{v},\mathbf{I}_j^{(2)})=1$) if the pixel at location $\boldsymbol{\pi}_u$ ({\it resp. $\boldsymbol{\pi}_v$}) in image $\mathbf{I}_i^{(1)}$ ({\it resp.} $\mathbf{I}_j^{(2)}$) is encoded by codeword $\mathbf{u}$ ({\it resp.} $\mathbf{v}$); otherwise, 0.

%Fig. \ref{fig:basis-function} illustrates the idea behind the co-occurrence statistics for {\it re-id}. Given a pair of images from different views, the co-occurrence statistics at the same locations, \ie $p(\mathbf{u},\mathbf{v}|\mathbf{h})$, will contribute to re-identification based on the weights $p(y_1=y_2|\mathbf{u},\mathbf{v})$ for the co-occurrence. The learned classifier $\mathbf{w}$ and the basis function in Eq. \ref{eqn:image_matching} can be interpreted as $p(y_1=y_2|\mathbf{u},\mathbf{v})$ and $p(\mathbf{u},\mathbf{v}|\mathbf{h})$, respectively. $p(\mathbf{h})$ is simply taken as a uniform distribution.

%The key insight of their model is that aspects of appearance that are transformed in predictable ways, due to the static camera view angles, can be statistically inferred through pairwise co-occurrence of visual words. 
%The computational cost of the descriptor, however, is very high.

%Inspired by \cite{zhang_eccv14}, here we propose a binary image descriptor, and further define our basis functions based on these descriptors. 

\subsubsection{Multi-Shot Visual Word Co-occurrence Models}
By comparing the simplified model in Eq. \ref{eqn:latent} with Eq. \ref{eqn:probabilistic}, we can set
\begin{align}\label{eqn:puv}
p(\mathbf{u},\mathbf{v}|\mathbf{h},\mathbf{I}_i^{(1)},\mathbf{I}_j^{(2)})\stackrel{\Delta}{=}\max_{\boldsymbol{\pi}_u\in\Pi_u}\kappa(\boldsymbol{\pi}_u,\mathbf{h})\cdot \max_{\boldsymbol{\pi}_v\in\Pi_v}\kappa(\boldsymbol{\pi}_v,\mathbf{h}),
\end{align}
where $\max_{\boldsymbol{\pi}_u\in\Pi_u}\kappa(\boldsymbol{\pi}_u,\mathbf{h})$ and $\max_{\boldsymbol{\pi}_v\in\Pi_v}\kappa(\boldsymbol{\pi}_v,\mathbf{h})$ can be computed independently once for comparing similarities, making the calculation much more efficient. This model cannot, however, handle the multi-shot scenario directly.

\begin{figure}[t]
\begin{center}
\centerline{\includegraphics[width=0.8\columnwidth]{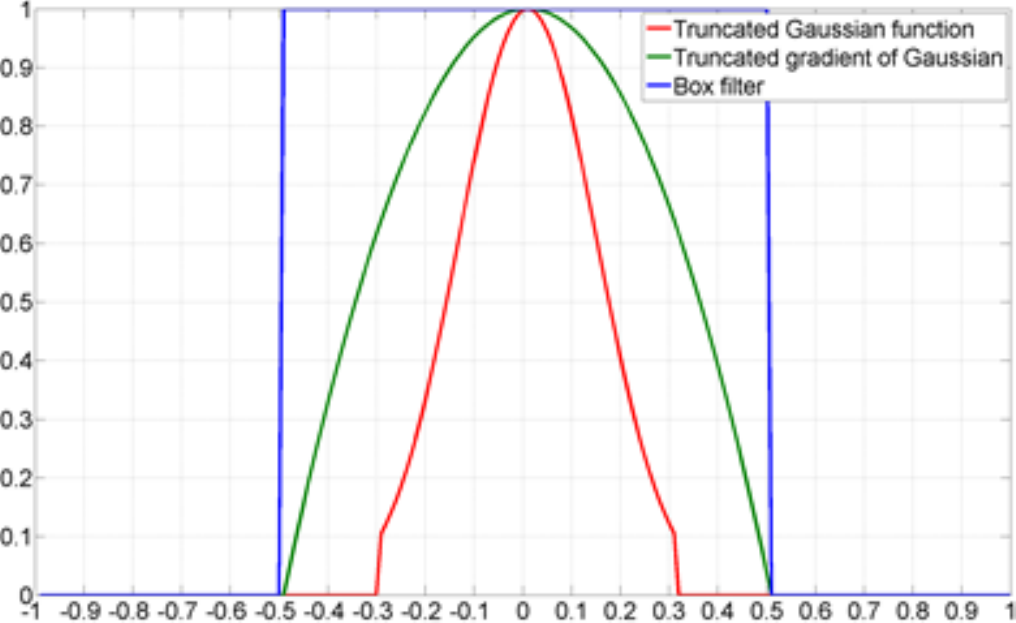}}
\caption{\footnotesize{Illustration of the three efficient spatial kernels for $\kappa$.}}\label{fig:sp_kernel}
\end{center}
%\vspace{-5mm}
\end{figure}

In this paper we extend the visual word co-occurrence model in \cite{ZhangECCV2014} to the multi-shot scenario, and propose three more efficient spatial kernel functions for $\kappa$. 

Let $\forall p,q, \mathcal{I}_q^{(p)}=\{\mathbf{I}_{m_q}^{(p)}\}_{m_q=1,\cdots,N_{q}^{(p)}}^{p=1,2}$ be the image set with the same image resolution for entity $q$ from view $p$, where $\forall m_q, \mathbf{I}_{m_q}^{(p)}$ denotes the $\left(m_{q}\right)^{th}$ image, and $\sigma$ be the scale of patches centered at these locations. Then we can define our $\phi(\mathbf{x}_{ij})$ as follows:
\begin{align}\label{eqn:basis_function}
\Big[\phi(\mathbf{x}_{ij})\Big]_{uv}=\sum_{\mathbf{h}\in\Pi}\Big[\psi(\mathcal{I}_{i}^{(1)},\mathbf{h},\sigma)\Big]_u\cdot\Big[\psi(\mathcal{I}_{j}^{(2)},\mathbf{h},\sigma)\Big]_v,
\end{align}
%\begin{align}\label{eqn:basis_function}
%& \phi_{st}(\mathbf{x}_{ij})=\sum_{m_1=1}^{N_{1i}}\sum_{m_2=1}^{N_{2j}}\sum_{\mathbf{h}\in\mathcal{H}}\frac{\psi_s(I_{m_1}^1,\mathbf{h},\sigma)}{N_{1i}}\cdot\frac{\psi_t(I_{m_2}^2,\mathbf{h},\sigma)}{N_{2j}}\nonumber\\
%&=\sum_{\mathbf{h}\in\mathcal{H}}{\Bigg[\sum_{m_1=1}^{N_{1i}}\frac{\psi_s(I_{m_1}^1,\mathbf{h},\sigma)}{N_{1i}}\Bigg]\cdot\Bigg[\sum_{m_2=1}^{N_{2j}}\frac{\psi_t(I_{m_2}^2,\mathbf{h},\sigma)}{N_{2j}}\Bigg]},
%\end{align}
where $\psi$ denotes the multi-shot visual word descriptor at location $\mathbf{h}$ for each entity, and $[\cdot]_u$ (\resp $[\cdot]_v$) denotes the entry in the vector for $\mathbf{u}$ (\resp $\mathbf{v}$) at location $\mathbf{h}$. 

%Next, we will define $\psi_s(\mathbf{x}_{m_i}^1,\mathbf{h},\sigma)$ and $\psi_t(\mathbf{x}_{m_j}^2,\mathbf{h},\sigma)$, so that they are tolerant to spatial distortion of visual word, and also can be used for both single-shot and multi-shot learning. Here, we will take $\psi_s(\mathbf{x}_{m_i}^1,\mathbf{h},\sigma)$ as an example to explain the definition. $\psi_t(\mathbf{x}_{m_j}^2,\mathbf{h},\sigma)$ can be defined in a similar way accordingly.

%We introduce a {\em spatial kernel} $\kappa$ to measure the spatial similarity between a visual word and the patch center location. 
Next, we will take $[\psi(\mathcal{I}_{i}^{(1)},\mathbf{h},\sigma)]_u$ as an example to explain its definition, and accordingly $[\psi(\mathcal{I}_{j}^{(2)},\mathbf{h},\sigma)]_v$ can be defined similarly. Letting $\Pi_u(\mathbf{I}_{m_i}^{(1)})$ be the set of locations where pixels are encoded by visual word $\mathbf{u}$ in image $\mathbf{I}_{m_i}^{(1)}$, based on Eq. \ref{eqn:puv} we define $[\psi(\mathcal{I}_{i}^{(1)},\mathbf{h},\sigma)]_u$ as follows:
\begin{equation}\label{eqn:k}
\Big[\psi(\mathcal{I}_{i}^{(1)},\mathbf{h},\sigma)\Big]_u=\frac{1}{|\mathcal{I}_{i}^{(1)}|}\sum_{\mathbf{I}_{m_i}^{(1)}\in\mathcal{I}_{i}^{(1)}}\max_{\boldsymbol{\pi}_u\in\Pi_u\left(\mathbf{I}_{m_i}^{(1)}\right)}\kappa\left(\boldsymbol{\pi}_u, \mathbf{h};\sigma\right),
\end{equation}
where $|\cdot|$ denotes the cardinality of a set, \ie the number of images for person entity $i$ from view 1, and $\sigma$ is the spatial kernel parameter controlling the locality. For multi-shot learning, we take each sequence as a collection of independent images, and utilize the average to represent the entity. Even though we use such simple representation (\eg totally ignoring the temporal relations between images), it turns out that our method can outperform the current state-of-the-art significantly for the multi-shot scenario, as we will demonstrate in Section \ref{ssec:ms}.

The choices for the spatial kernel $\kappa$ in Eq. \ref{eqn:k} are quite flexible. To account for computational efficiency, here we list three choices, \ie (1) truncated Gaussian filters ($\kappa_1$), (2) truncated gradient of Gaussian filters ($\kappa_2$), and (3) box filters \cite{rau1997efficient} ($\kappa_3$). Their definitions are shown below:
\begin{align}
\label{eqn:kappa_1}
& \kappa_1=\left\{
\begin{array}{ll}
\exp\left\{-\frac{\mathbf{dist}(\boldsymbol{\pi}_s, \mathbf{h})}{\sigma_1}\right\}, & \mbox{if} \; \mathbf{dist}(\boldsymbol{\pi}_s, \mathbf{h})\leq\alpha\\
0, & \mbox{otherwise}.
\end{array}
\right.\\
\label{eqn:kappa_2}
& \kappa_2=\max\left\{0, 1-\frac{1}{\sigma_2}\cdot\mathbf{dist}(\boldsymbol{\pi}_s, \mathbf{h})\right\}, \\
\label{eqn:kappa_3}
& \kappa_3=\left\{
\begin{array}{ll}
1, & \mbox{if} \; \mathbf{dist}(\boldsymbol{\pi}_s, \mathbf{h})\leq \sigma_3, \\
0, & \mbox{otherwise}.
\end{array}
\right.
\end{align}
where $\mathbf{dist}(\cdot,\cdot)$ denotes a distance function, $\sigma_1,\sigma_2,\sigma_3$ denote the corresponding scale parameters in the functions, and $\alpha\geq0$ in Eq. \ref{eqn:kappa_1} is a predefined thresholding parameter. These three functions are illustrated in Fig. \ref{fig:sp_kernel}, where the distance function is the Euclidean distance. Compared with the Gaussian function that is used in \cite{ZhangECCV2014}, these three functions produce much sparser features, making the computation more efficient.
%From the view of function smoothness, $\kappa_1$ is the best, but from the view of feature sparsity, both $\kappa_2$ and $\kappa_3$ are better than $\kappa_1$. As we see in Section \ref{sec:exp}, function smoothness and feature sparsity have strong impact on matching rate and computational efficiency of our method, respectively.

\begin{figure*}[t]
\begin{center}
\centerline{\includegraphics[width=0.9\linewidth]{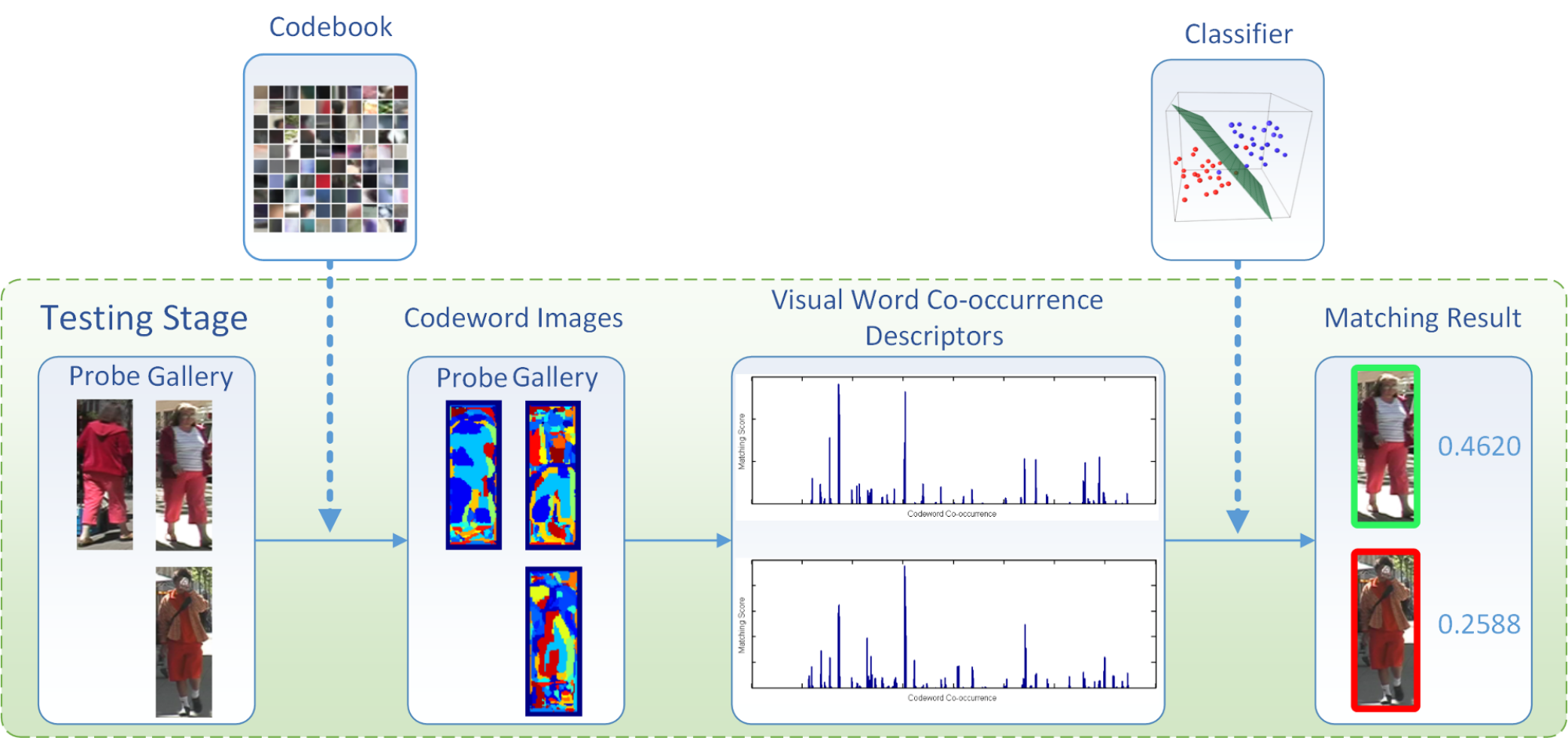}}
\caption{\footnotesize{The pipeline of our method, where ``codebook'' and ``classifier'' are learned using training data, and each color in the codeword images denotes a codeword.}}\label{fig:Pipeline}
\end{center}
%\vspace{-5mm}
\end{figure*}

\subsection{Structured Learning of Similarity Models}\label{ssec:sl}
Now we come to the other question of as to how we learn the weight vector $\mathbf{w}$ in Eq. \ref{eqn:test}. %We use structured learning.

We denote the training entity set as $\mathcal{X}=\left\{\mathbf{x}_q^p\right\}_{q=1,\cdots,N_p}^{p=1,2}$, where $\forall p, \forall q, \mathbf{x}_q^p$ denotes the $q^{th}$ person entity from camera view $p$. We refer to view 1 as the {\em probe} set, and view 2 as the {\em gallery} set. Also, we denote $\mathbf{y}=\{y_{ij}\}_{i,j\geq1}$ as the ground-truth bipartite graph structure, and $y_{ij}=1$ if $\mathbf{x}_i^{(1)}$ and $\mathbf{x}_j^{(2)}$ are the same; otherwise, $y_{ij}=0$. Then our method in training can be formulated as the following structured learning problem:
\begin{align}\label{eqn:image_matching}
\min_{\mathbf{w}, \xi} & \; \frac{1}{2}\|\mathbf{w}\|_2^2+C\xi \\
\mbox{s.t.} & \; \forall \bar{\mathbf{y}}\in\mathcal{Y},  \mathbf{w}^Tf(\mathcal{X},\mathbf{y})\geq\mathbf{w}^Tf(\mathcal{X},\bar{\mathbf{y}})+\Delta(\mathbf{y},\bar{\mathbf{y}})-\xi, \nonumber \\
& \; \xi\geq0, \nonumber
\end{align}
where $\mathbf{w}$ is the weight vector, $\bar{\mathbf{y}}\in\mathcal{Y}$ denotes a predicted bipartite graph structure, $f(\mathcal{X},\mathbf{y})=\sum_{i,j}y_{ij}\phi(\mathbf{x}_{ij})$ (\resp $f(\mathcal{X},\bar{\mathbf{y}})=\sum_{i,j}\bar{y}_{ij}\phi(\mathbf{x}_{ij})$) denotes the basis function under the ground-truth (\resp predicted) graph structure, $\Delta(\mathbf{y},\bar{\mathbf{y}})=\sum_{i,j}|y_{ij}-\bar{y}_{ij}|$ denotes the loss between the two structures, $C\geq0$ is a predefined regularization constant, and $\|\cdot\|_2$ denotes the $\ell_2$ norm of a vector. Here the constraint is enforcing the structured matching score of the ground-truth structure to be the highest among all possible matching structures in $\mathcal{Y}$. In order to adapt the definition of $\mathcal{Y}$ in Eq. \ref{eqn:y} to the ground-truth matching structure $\mathbf{y}$, we can simply set $r_i=\max_i\sum_jy_{ij}$, and substitute this value into Eq. \ref{eqn:y} to construct the feasible solution space $\mathcal{Y}$. Same as the structured matching in testing, in training we also utilize a priori knowledge on the correct matching structures to reduce the chance of mismatching.

\begin{algorithm}[t]
\SetAlgoLined
\SetKwInOut{Input}{Input}\SetKwInOut{Output}{Output}
\Input{training entity set $\mathcal{X}$, ground-truth matching structure $\mathbf{y}$, predefined regularization parameter $C\geq0$}
\Output{$\mathbf{w}$}
\BlankLine
Construct the feasible solution space $\mathcal{Y}$; \\
Randomly sample a subset of matching structures $\bar{\mathcal{Y}}\subset\mathcal{Y}$;\\
\Repeat{Converge}
{
$\mathbf{w}\leftarrow\mbox{RankSVM\_Solver}(\mathcal{X}, \mathbf{y}, \bar{\mathcal{Y}}, C)$;\\
$\mathbf{y}^*\leftarrow\argmax_{\bar{\mathbf{y}}\in\mathcal{Y}}\mathbf{w}^Tf(\mathcal{X}, \bar{\mathbf{y}}) + \Delta(\mathbf{y},\bar{\mathbf{y}})$;\\
$\bar{\mathcal{Y}}\leftarrow\bar{\mathcal{Y}}\bigcup\mathbf{y}^*$;
}
\Return $\mathbf{w}$\;
\caption{Structured learning of PRISM}\label{alg:image_matching}
\end{algorithm}

In principle we can solve Eq. \ref{eqn:image_matching} using 1-slack structural SVMs \cite{Joachims:2009:CTS:1612990.1613006}. We list the cutting-plane algorithm for training PRISM in Alg. \ref{alg:image_matching}. The basic idea here is to select most violated matching structure $\mathbf{y}^*$ from the feasible set $\mathcal{Y}$ in each iteration, and add it into the current feasible set $\bar{\mathcal{Y}}$, and resolve Eq. \ref{eqn:image_matching} using $\bar{\mathcal{Y}}$. In this way, the solution searching space is dependent on $\bar{\mathcal{Y}}$ rather than $\mathcal{Y}$. In each iteration, we can simply adopt RankSVM solver \cite{Lee:2014:LLR:2645408.2645415} to find a suitable $\mathbf{w}$. For inference, since we have $\Delta(\mathbf{y},\bar{\mathbf{y}})=\sum_{i,j}|y_{ij}-\bar{y}_{ij}|=\sum_{i,j}(y_{ij}-\bar{y}_{ij})^2$ (because of $\forall y_{ij}\in\{0, 1\}, \forall\bar{y}_{ij}\in\{0, 1\}$), we indeed solve a binary quadratic problem, which can be efficiently solved using the similar thresholding trick for inference in testing.

Note that in order to speed up the learning we can alternatively adopt large-scale linear RankSVMs \cite{Lee:2014:LLR:2645408.2645415} (or even linear SVMs \cite{REF08a} as we did in \cite{ZhangECCV2014}) with a large amount of randomly sampled matching structures from $\mathcal{Y}$ to approximately solve Eq. \ref{eqn:image_matching}. This trick has been widely used in many large-scale training methods (\eg \cite{Tong:2002:SVM:944790.944793}) and demonstrated its effectiveness and efficiency without notable performance loss. Similarly, in our re-id cases we implement both learning strategies and have found that the performance loss is marginal.

\section{Implementation}\label{sec:implementation}
We illustrate the schematics of our method in Fig. \ref{fig:Pipeline}. At training stage, we extract low-level feature vectors from randomly sampled patches in training images, and then cluster them into codewords to form a codebook, which is used to encode every image into a codeword image. Each pixel in a codeword image represents the centroid of a patch that has been mapped to a codeword. Further, a visual word co-occurrence model (descriptor) is calculated for every pair of gallery and probe images, and the descriptors from training data are utilized to train our classifier using Eq. \ref{eqn:image_matching}. We perform re-id on the test data using Eq. \ref{eqn:test}.

Specifically, we extract a 672-dim Color+SIFT\footnote{We downloaded the code from \url{https://github.com/Robert0812/salience_match}.} \cite{Zhao_ICCV13_salience} feature vector from each 5$\times$5 pixel patch in images, and utilize K-Means to generate the visual codebooks based on about $3\times10^4$ randomly selected Color+SIFT features per camera view. Then every Color+SIFT feature is quantized into one of these visual words based on minimum Euclidean distance. The number of visual words per view is set by cross-validation.

%We first resize all images into $128\times48$ pixels. Then we take every $2\times2$ pixel patch of images in the HSV color space, and concatenate $3\times2\times2=12$ entries into a vector as our 12-dim low level features, since color features have been demonstrated very useful for \reid \cite{yang_eccv14}. 
%However, {\em no one has published any result on \reid using such simple features, to our best knowledge.} With the help of this HSV features, our method has great potential for real-time applications (see Section \ref{ssec:ct} for details).

\begin{figure*}[t]
\begin{center}
\centerline{\includegraphics[width=0.8\linewidth]{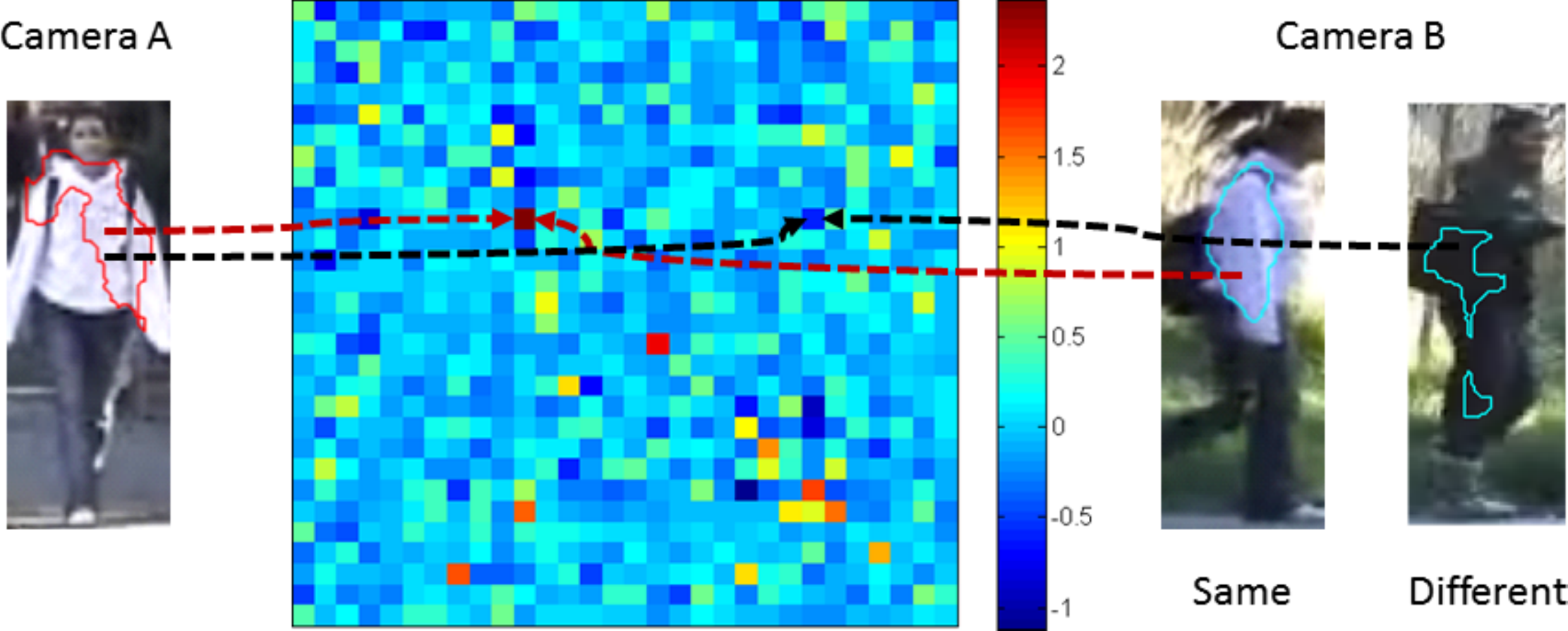}}
\caption{\footnotesize{The interpretation of our learned model parameter $\mathbf{w}$ in Eq. \ref{eqn:image_matching}. The enclosed regions denote the pixels encoded by the same visual words, as used in Fig. \ref{fig:intuition}. The learned weight for the visual word pair ``white'' and ``light-blue'' from the two camera views has a positive value, contributing to identifying the same person. On the other hand, the learned weight for the visual word pair ``white'' and ``black'' is negative, which contributes to identifying different persons.}}\label{fig:interpret}
\end{center}
\vspace{-5mm}
\end{figure*}

We employ the chessboard distance for Eq. \ref{eqn:kappa_1}, \ref{eqn:kappa_2} and \ref{eqn:kappa_3}, consider every pixel location as $\mathbf{h}$, and set the scale parameter $\sigma$ by cross-validation for the spatial kernel $\kappa$. Similarly the regularization parameter $C$ in Eq. \ref{eqn:image_matching} is set by cross-validation. 
%We prefer using the ground-truth graph as illustrated in Fig. \ref{fig:y-ex}(b), which includes all the positive and negative matching pairs for training. However, in order to overcome memory size limit, if necessary, we also randomly create the ground-truth graph for training which includes all the correct entity matches, as illustrated in Fig. \ref{fig:y-ex}(c).

During testing, for performance measure we utilize a standard metric for re-id, namely, Cumulative Match Characteristic (CMC) curve, which displays an algorithm's recognition rate as a function of rank. For instance, a recognition rate at rank-$r$ on the CMC curve denotes what proportion of queries are correctly matched to a corresponding gallery entity at rank-$r$ or better. Therefore, we set $\forall i, r_i=r$ in Eq. \ref{eqn:test}, and solve the optimization problem.

Note that we can further save on computational time for prediction during testing. This follows from the fact that we do not need the exact solution for Eq. \ref{eqn:test}, as long as we can use the current solution to find the ranks of entities in the gallery for each probe, to determine the top matches. Therefore, we ask the linear programming solver to run for 10 iterations at most in our experiments.

%Based on the implementation and notations above, for comparing the similarity of a pair of entities $\mathbf{x}_{ij}$, our computational complexity during testing can be mainly divided into 4 parts: (1) $O(12\cdot|\mathcal{H}|\cdot(N_uN_{1i}+N_vN_{2j}))$ for HSV feature quantization, where $|\mathcal{H}|$ denotes the number of pixels in images, (2) , (3) $O(|\mathcal{H}|\cdot(N_uN_{1i}+N_vN_{2j}+N_uN_v))$ for calculating the basis function, (4) $O(N_uN_v)$ for calculating the entity similarity.

\section{Experiments}\label{sec:exp}
We test our method on three benchmark datasets, \ie VIPeR \cite{Gray_PETS07_VIPER}, CUHK Campus \cite{Zhao_ICCV13_salience}, and  iLIDS-VID \cite{wang_eccv14}, for both single-shot and multi-shot scenarios. We do not re-implement comparative methods.
%, because we are not sure whether our own implementation will reproduce the similar results as published in the papers. 
Instead, we try to cite numbers/figures of comparative methods either from released codes or from the original papers as accurately as possible (\ie for methods LAFT \cite{conf/cvpr/LiW13} and LDM \cite{2165365} in Table \ref{tab:ss} and Table \ref{tab:ms}, respectively), if necessary. Also, we compare our method against currently known state-of-the-art on these datasets. Our experimental results are reported as the average over 3 trials.

We denote the three derivatives of our method based on different spatial kernels as (1) {\sc PRISM-\rmnum{1}} for using $\kappa_1$ in Eq. \ref{eqn:kappa_1}, (2) {\sc PRISM-\rmnum{2}} for using $\kappa_2$ in Eq. \ref{eqn:kappa_2}, and (3) {\sc PRISM-\rmnum{3}} for using $\kappa_3$ in Eq. \ref{eqn:kappa_3}, respectively.
%Our code will be published upon acceptance of the paper.

\subsection{Datasets and Experimental Settings}
VIPeR \cite{Gray_PETS07_VIPER} consists of 632 entities captured in two different camera views, denoted by CAM-A and CAM-B, respectively. Each image is normalized to 128$\times$48 pixels. We follow the experimental set up described in \cite{Zhao_ICCV13_salience}. The dataset is split in half randomly, one partition for training and the other for testing. Samples from CAM-A and CAM-B form the probe and gallery sets, respectively.

CUHK Campus \cite{conf/cvpr/LiW13,Zhao_ICCV13_salience} consists of 1816 people captured from five different camera pairs, labeled from P1 to P5 and denoted as CAM-1 and CAM-2 per camera pair which form the probe and gallery sets, respectively. Each camera view has 2 images per entity, and each image contains 160$\times$60 pixels. We follow the experimental settings in \cite{conf/cvpr/LiW13,Zhao_ICCV13_salience}, and use only images captured from P1. We randomly select 485 individuals from the dataset for training, and use the rest 486 individuals for testing.

iLIDS-VID \cite{wang_eccv14} is a new \reid dataset created based on two non-overlapping camera views from the i-LIDS Multiple-Camera Tracking Scenario (MCTS) \cite{iLIDS}. For single-shot learning, there are 300 image pairs for 300 randomly selected people with image size equal to $128\times64$ pixels. For multi-shot learning, there are 300 pairs of image sequences for the 300 people. The length of each image sequence varies from 23 to 192 frames with average of 73. Following \cite{wang_eccv14}, we randomly select 150 people as training data, and use the rest 150 people as testing data. The data from the first and second camera views forms the probe and gallery sets, respectively.

\begin{figure*}[t]
\begin{minipage}[b]{0.49\linewidth}
 \begin{center}
 \centerline{\includegraphics[width=0.95\columnwidth]{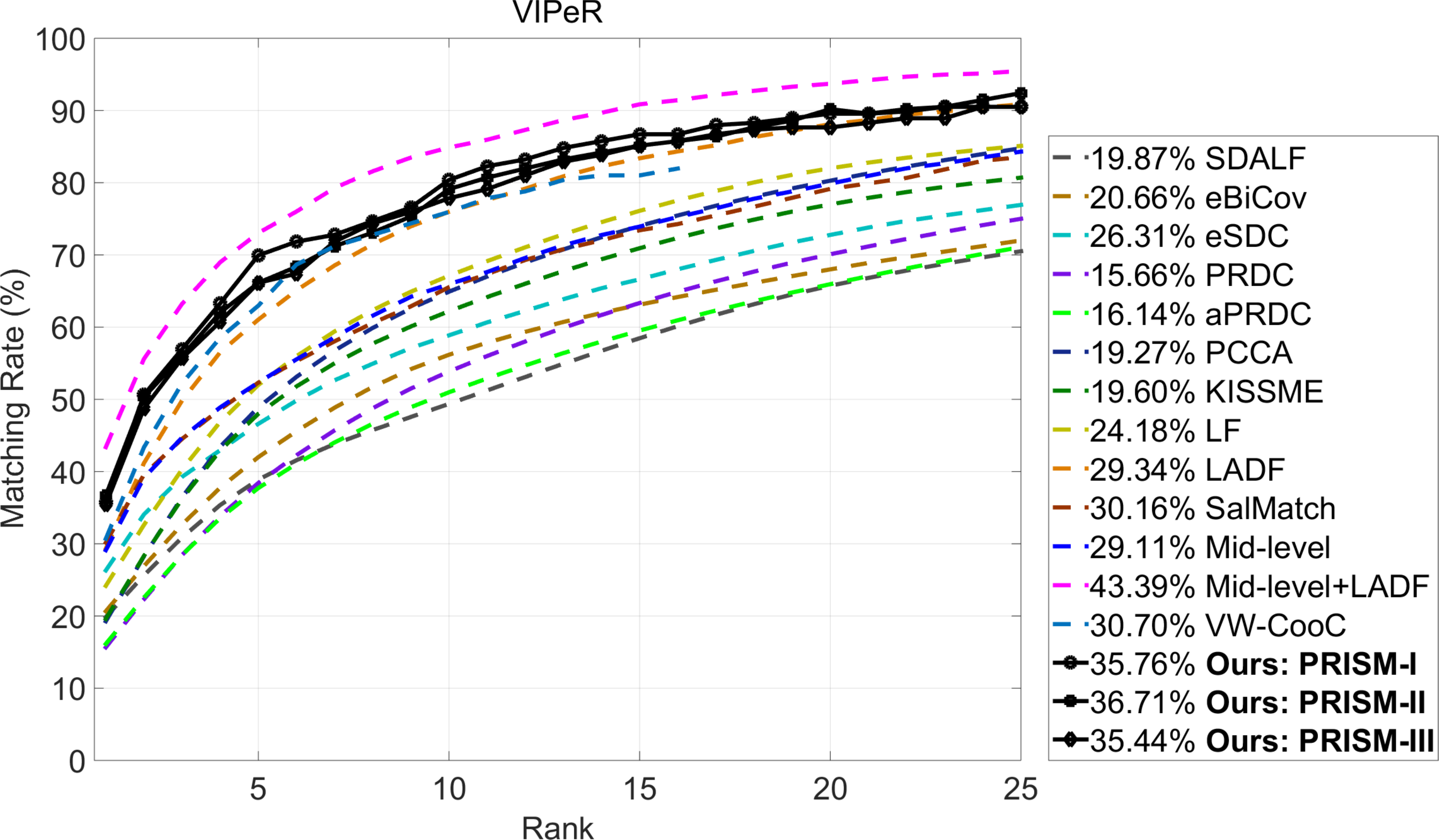}}
 \centerline{\footnotesize{(a)}}
 \end{center}
% \vspace*{-5mm}
\end{minipage}
\begin{minipage}[b]{0.49\linewidth}
\begin{center}
\centerline{\includegraphics[width=0.95\columnwidth]{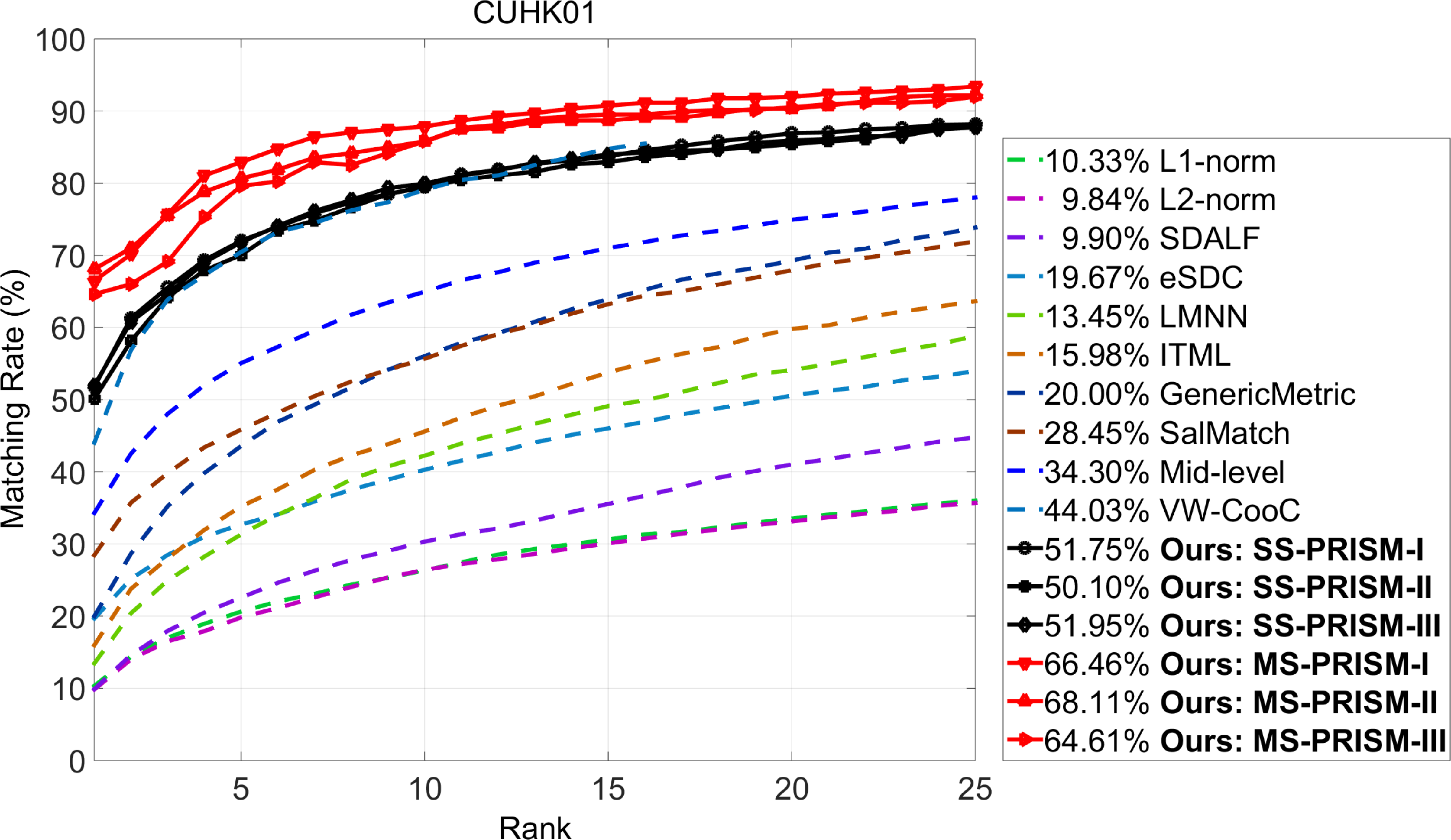}}
 \centerline{\footnotesize{(b)}}
\end{center} 
\end{minipage}
\vspace{-3mm}
\caption{\footnotesize{CMC curve comparison on \textbf{(a)} VIPeR and \textbf{(b)} CUHK01, where ``SS'' and ``MS'' denote the single-shot and multi-shot, respectively. Notice that except our results, the rest are copied from \cite{zhao2014learning}.} }\label{fig:single-shot}
%\vspace{-3mm}
\end{figure*}

%\begin{figure*}[t]
%\begin{center}
%\centerline{\includegraphics[width=\linewidth]{matching_examples.pdf}}
%\caption{\footnotesize{Examples of codeword co-occurrence with relatively high positive/negative weights in the learned weighting matrix. Same as Fig. \ref{fig:intuition}, in each row the regions enclosed by red (or cyan) color indicate that the codeword per pixel location in these regions is the same. This figure is best viewed in color.}}\label{fig:matching_examples}
%\end{center}
%%\vspace{-5mm}
%\end{figure*}

\subsection{Model Interpretation}
We start by interpreting our learned model parameters $\mathbf{w}$ in Eq. \ref{eqn:image_matching}. We show a typical learned $\mathbf{w}$ matrix in Fig. \ref{fig:interpret} with 30 visual words per camera view. Recall that $\mathbf{w}\stackrel{\Delta}{=}p(\bar{y}_{ij}=1|\mathbf{u},\mathbf{v})$ denotes how likely two images come from a same person according to the visual word pairs, and our spatial kernel $\kappa$ always returns non-negatives indicating the spatial distances between visual word pairs in two images from two camera views. As we see in Fig. \ref{fig:interpret}, by comparing the associated learned weights, ``white'' color in camera $A$ is likely to be transfered into ``light-blue'' color (with higher positive weight), but very unlikely to be transfered into ``black'' color (with lower negative weight) in camera $B$. Therefore, when comparing two images from camera $A$ and $B$, respectively, if within the same local regions the ``white'' and ``light-blue'' visual word pair from the two images occurs, it will contribute to identifying the same person; on the other hand, if ``white'' and ``black'' co-occur within the same local regions in the images, it will contribute to identifying different persons.

\subsection{Single-Shot Learning}

\begin{savenotes}
\begin{table}[t]\centering
\caption{\footnotesize{Matching rate comparison (\%) for single-shot learning, where ``-'' denotes no result reported for the method.}}\label{tab:ss}
\setlength\tabcolsep{3pt}
\begin{tabular}{|l|llllll|}
\hline
Rank $r=$ & 1 & 5 & 10 & 15 & 20 & 25 \\
\hline\hline

& \multicolumn{6}{c|}{VIPeR} \\
\hline
SCNCD \cite{yang_eccv14}    & 20.7 & 47.2  &  60.6  &  68.8  &  75.1  &  79.1  \\
LADF \cite{conf/cvpr/LiCLHCS13}    & 29.3  & 61.0  &  76.0  &  83.4  &  88.1  &  90.9 \\
Mid-level filters \cite{zhao2014learning}    & 29.1  & 52.3  &  65.9  &  73.9  &  79.9  &  84.3 \\
Mid-level filters+LADF \cite{zhao2014learning} & {\bf 43.4} & {\bf 73.0} & {\bf 84.9} & {\bf 90.9} & 93.7 & {\bf 95.5} \\
VW-CooC \cite{ZhangECCV2014} & 30.7 & 63.0 & 76.0 & 81.0 & - & - \\
RQDA \cite{liao2014joint} & 34.7 & 65.4 & 78.6 & - & 89.6 & - \\
Semantic (super. single) \cite{zheng2015query} & 31.1 & 68.6 & 82.8 & - & {\bf 94.9} & - \\
Polynomial kernel \cite{chen2015similarity} & 36.8 & 70.4 & 83.7 & - & 91.7 & - \\
\hline
QALF \cite{liao2014joint} & 30.2 & 51.6 & 62.4 & - & 73.8 & - \\
Semantic (super. fusion) \cite{zheng2015query} & 41.6 & 71.9 & 86.2 & - & 95.1 & - \\
SCNCD$_{final}$(ImgF) \cite{yang_eccv14}\footnote{ImgF: image-foreground feature representations}    &  37.8 &  68.5 &  81.2  &  {\bf 87.0}  &  90.4  &  {\bf 92.7}  \\
Ensemble Color Model \cite{LiuWACV2015} & 38.9 & 67.8 & 78.4 & - & 88.9 & - \\
Metric ensembles \cite{2015arXiv150301543P} & {\bf 45.9} & {\bf 77.5} & {\bf 88.9} & - & {\bf 95.8} & - \\
Kernel ensembles-{\sc \rmnum{1}} \cite{xiong2014person} & 35.1 & 68.2 & 81.3 & - & 91.1 & - \\
Kernel ensembles-{\sc \rmnum{2}} \cite{xiong2014person} & 36.1 & 68.7 & 80.1 & - & 85.6 & - \\
\hline
\textbf{Ours:} {\sc PRISM-\rmnum{1}} & 35.8 & {\bf 69.9} & {\bf 80.4} & {\bf 86.7} & 89.6 & 90.5 \\
\textbf{Ours:} {\sc PRISM-\rmnum{2}} & {\bf 36.7} & 66.1 & 79.1 & 85.1 & {\bf 90.2} & {\bf 92.4} \\
\textbf{Ours:} {\sc PRISM-\rmnum{3}} & 35.4 & 66.1 & 77.9 & 85.1 & 87.7 & 90.5 \\
\hline\hline
& \multicolumn{6}{c|}{CUHK01} \\
\hline
LAFT \cite{conf/cvpr/LiW13} & 25.8 & 55.0 & 66.7 & 73.8 & {\bf 79.0} & {\bf 83.0} \\
Mid-level filters \cite{zhao2014learning}    & 34.3 & 55.1 &  65.0  &  71.0  &  74.9  &  78.0 \\
VW-CooC \cite{ZhangECCV2014} & {\bf 44.0} & {\bf 70.5} & {\bf 79.1} & {\bf 84.8} & - & - \\
Semantic (super. single) \cite{zheng2015query} & 32.7 & 51.2 & 64.4 & - & 76.3 & - \\
\hline
Semantic (super. fusion) \cite{zheng2015query} & 31.5 & 52.5 & 65.8 & - & 77.6 & - \\
Metric ensembles \cite{2015arXiv150301543P} & {\bf 53.4} & {\bf 76.4} & {\bf 84.4} & - & {\bf 90.5} & - \\
\hline
\textbf{Ours:} {\sc PRISM-\rmnum{1}} & 51.8 & {\bf 72.0} & 79.5 & 83.7 & {\bf 86.9} & {\bf 88.2} \\
\textbf{Ours:} {\sc PRISM-\rmnum{2}} & 50.1 & 70.1 & 79.4 & 82.9 & 85.4 & 87.8 \\
\textbf{Ours:} {\sc PRISM-\rmnum{3}} & {\bf 52.0} & 71.8 & {\bf 79.9} & {\bf 84.0} & 85.9 & 87.8 \\
\hline\hline
& \multicolumn{6}{c|}{iLIDS-VID} \\
\hline
Colour\&LBP \cite{DBLP:conf/eccv/HirzerRKB12}+RSVM & 9.1 & 22.6 & 33.2 & 45.5 & - & - \\
SS-SDALF \cite{Farenzena_CVPR10_SDALF} & 5.1 & 14.9 & 20.7 & 31.3  & - & - \\
Salience \cite{zhao2013unsupervised} & {\bf 10.2} & {\bf 24.8} & {\bf 35.5} & {\bf 52.9}  & - & - \\
\hline
\textbf{Ours:} {\sc PRISM-\rmnum{1}} & {\bf 22.0} & {\bf 43.3} & 52.0 & {\bf 62.7} & {\bf 73.3} & {\bf 77.3} \\
\textbf{Ours:} {\sc PRISM-\rmnum{2}} & 20.0 & 39.3 & {\bf 52.7} & 60.0 & 70.0 & 76.7 \\
\textbf{Ours:} {\sc PRISM-\rmnum{3}} & 16.7 & 36.7 & 52.0 & 56.7 & 67.3 & 74.7 \\
\hline
\end{tabular}
\end{table}
\end{savenotes}

\begin{figure*}[t]
\begin{minipage}[b]{0.33\linewidth}
 \begin{center}
 \centerline{\includegraphics[width=0.95\columnwidth]{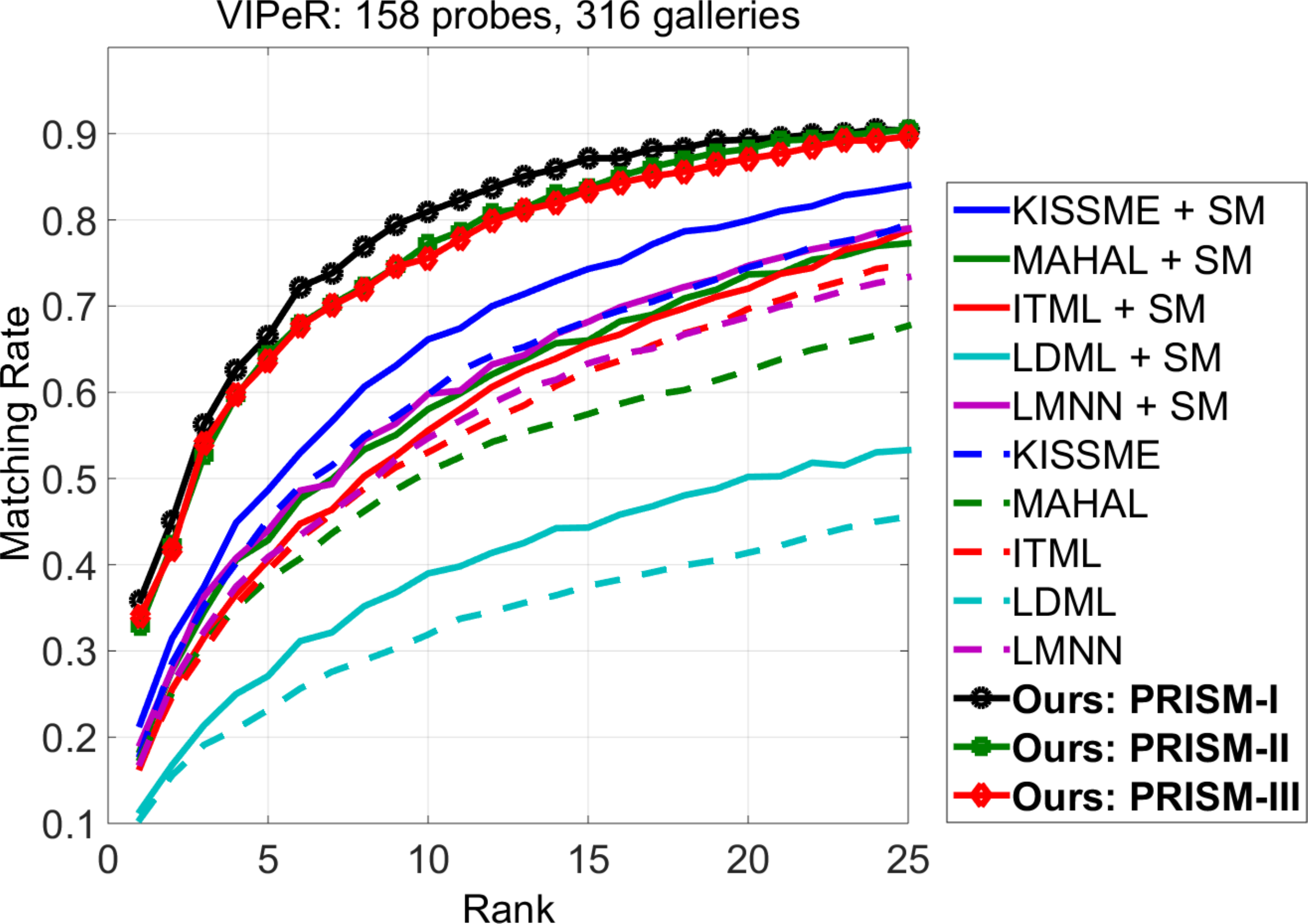}}
% \centerline{\footnotesize{(a)}}
 \end{center}
% \vspace*{-5mm}
\end{minipage}
\begin{minipage}[b]{0.33\linewidth}
\begin{center}
\centerline{\includegraphics[width=0.95\columnwidth]{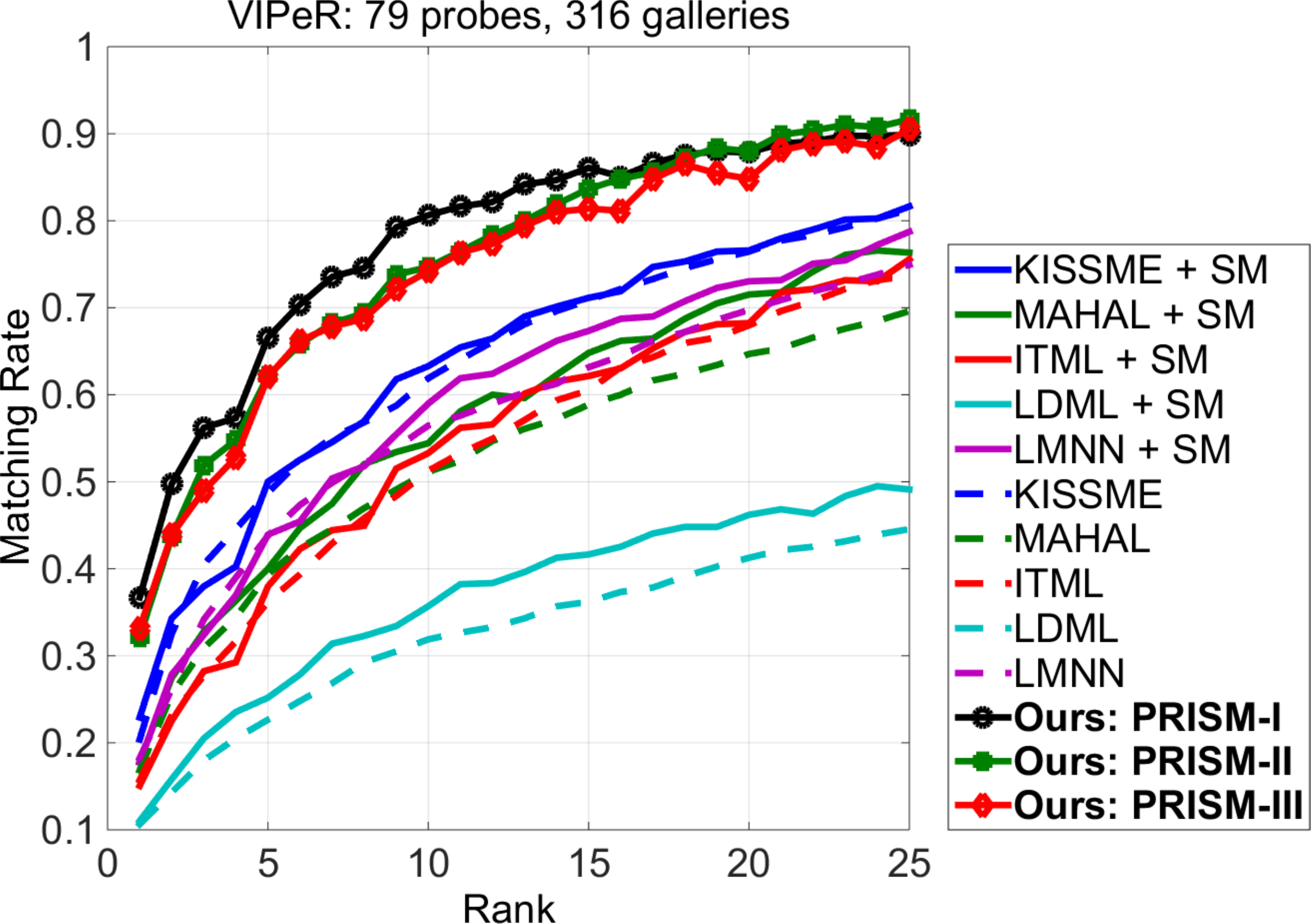}}
% \centerline{\footnotesize{(b)}}
\end{center} 
\end{minipage}
\begin{minipage}[b]{0.33\linewidth}
\begin{center}
\centerline{\includegraphics[width=0.95\columnwidth]{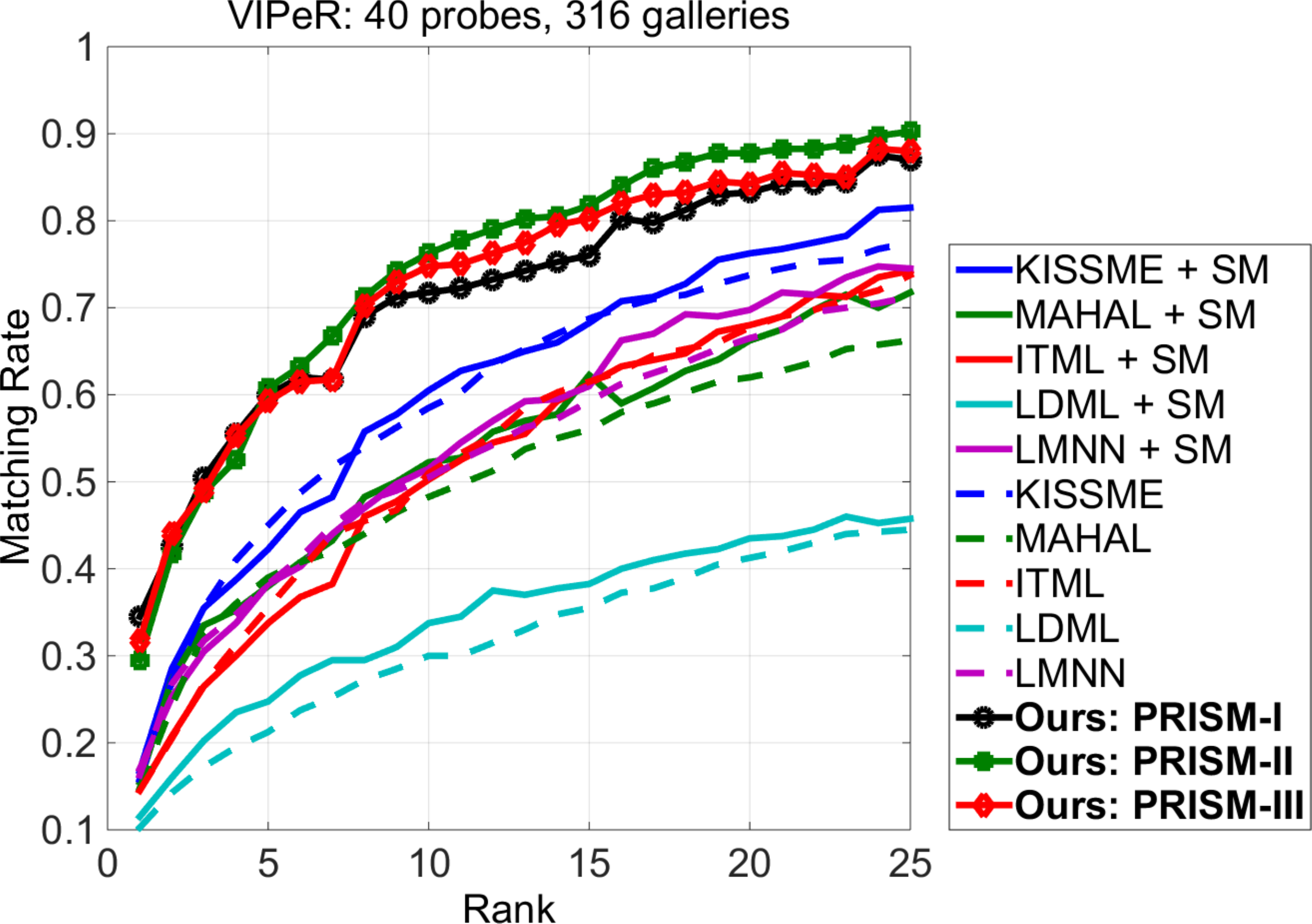}}
% \centerline{\footnotesize{(c)}}
\end{center} 
\end{minipage}
\vspace{-6.5mm}
\caption{\footnotesize{CMC curve comparison on VIPeR using different numbers of entities in the probe set for robust re-id.} }\label{fig:open-world}
%\vspace{-3mm}
\end{figure*}

Now we discuss our results for single-shot learning (see the definition in Section \ref{ssec:rw}). Table \ref{tab:ss} lists our comparison results on the three datasets, where the numbers are the matching rates over different ranks on the CMC curves. 

Here we divide comparative methods into 2 subcategories: non-fusion based and fusion based methods. Fusion based methods aim to combine multiple features/metrics to improve the matching performance, while non-fusion methods perform recognition using single type of features or a metric. Overall, fusion based methods achieve better performance than non-fusion based methods (including ours, which is always comparable). These methods, however, lack of clear interpretability of why the performance is better. Among non-fusion based methods, on VIPeR ``Mid-level filters+LADF'' from \cite{zhao2014learning} is the current best method, which utilized more discriminative mid-level filters as features with a powerful classifier, and ``SCNCD$_{final}$(ImgF)'' from \cite{yang_eccv14} is the second, which utilized only foreground features. Our results are comparable to both of them. However, PRISM always outperforms their original methods significantly when either the powerful classifier or the foreground information is not used. On CUHK01 and iLIDS-VID, PRISM performs the best. At rank-1, it outperforms \cite{ZhangECCV2014} and \cite{zhao2013unsupervised} by {\bf 8.0\%} and {\bf 11.8\%}, respectively. Some CMC curves of different methods on VIPeR and CUHK01 are compared in Fig. \ref{fig:single-shot}. Our current method only utilizes the visual word co-occurrence model. Integration of multiple features will be explored in our future work.

Compared with our previous work in \cite{ZhangECCV2014}, our improvement here mainly comes from the structured matching in testing by precluding the matches that are probably wrong (\ie reducing the feasible solution space). Clearly our method outperforms \cite{ZhangECCV2014} by {\bf 6.0\%} on VIPeR and {\bf 8.0\%} on CUHK01 at rank-1 rank in terms of matching rate.

%Fig. \ref{fig:matching_examples} illustrates some codeword co-occurrence examples with relatively high positive/negative weights in the learned weighting matrix. These examples strongly support our intuition of learning codeword co-occurrence based features in Section \ref{sec:intr}. 

%In summary, for single-shot learning, structured prediction can be utilized for improving the matching rate significantly, especially for the lower rank cases (\eg rank-1). Our PRISM works well on all the three datasets, which beats most of the current state of the art.

\subsection{Multi-Shot Learning}\label{ssec:ms}

\begin{savenotes}
\begin{table}[t]\centering
\caption{\footnotesize{Matching rate comparison (\%) for multi-shot learning, where ``-'' denotes no result reported for the method.}}\label{tab:ms}
\setlength\tabcolsep{3pt}
\begin{tabular}{|l|llllll|}
\hline
Rank $r=$ & 1 & 5 & 10 & 15 & 20 & 25 \\
\hline\hline
& \multicolumn{6}{c|}{CUHK01} \\
\hline
LAFT \cite{conf/cvpr/LiW13} & {\bf 31.4} & {\bf 58.0} & {\bf 68.3} & {\bf 74.0} & {\bf 79.0} & {\bf 83.0} \\
LDM \cite{2165365}    & 12.1  & 31.7  &  41.7  &  48.3  &  54.0  &  58.0 \\
\hline
\textbf{Ours:} {\sc PRISM-\rmnum{1}} & 66.5 & {\bf 82.9} & {\bf 87.9} & {\bf 90.7} & {\bf 92.0} & {\bf 93.4} \\
\textbf{Ours:} {\sc PRISM-\rmnum{2}} & {\bf 68.1} & 80.7 & 85.8 & 88.7 & 90.3 & 92.2 \\
\textbf{Ours:} {\sc PRISM-\rmnum{3}} & 64.6 & 79.6 & 85.8 & 89.5 & 90.5 & 92.0 \\
\hline\hline

& \multicolumn{6}{c|}{iLIDS-VID} \\
\hline
MS-SDALF \cite{Farenzena_CVPR10_SDALF} & 6.3 & 18.8 & 27.1 & - & 37.3 & - \\
MS-Colour\&LBP+RSVM  & 23.2  & 44.2  &  54.1  &  -  &  68.8  &  - \\
DVR \cite{wang_eccv14} & 23.3 & 42.4 & 55.3 & - & 68.4 & - \\
MS-SDALF+DVR & 26.7 & 49.3 & 61.0 & - & 71.6 & - \\
MS-Colour\&LBP+DVR & {\bf 34.5} & {\bf 56.7} & {\bf 67.5} & - & {\bf 77.5} & - \\
Salience \cite{zhao2013unsupervised}+DVR & 30.9 & 54.4 & 65.1 & - & 77.1 & - \\
\hline
\textbf{Ours:} {\sc PRISM-\rmnum{1}} & 60.7 & {\bf 86.7} & 89.3 & {\bf 94.7} & {\bf 96.0} & {\bf 96.7} \\
\textbf{Ours:} {\sc PRISM-\rmnum{2}} & {\bf 62.0} & 86.0 & {\bf 90.0} & {\bf 94.7} & {\bf 96.0} & {\bf 96.7} \\
\textbf{Ours:} {\sc PRISM-\rmnum{3}} & {\bf 62.0} & 86.0 & {\bf 90.0} & {\bf 94.7} & {\bf 96.0} & {\bf 96.7} \\
\hline
\end{tabular}
\end{table}
\end{savenotes}

\begin{savenotes}
\begin{table*}[t]\centering
\caption{\footnotesize{Matching accuracy comparison (\%) for robust re-id.}}\label{tab:open-world}
\setlength\tabcolsep{3pt}
\begin{tabular}{|l||llllll|llllll|llllll|}
\hline
& \multicolumn{6}{c|}{158 probes} & \multicolumn{6}{c|}{79 probes} & \multicolumn{6}{c|}{40 probes} \\
\hline
Rank $r=$ & 1 & 5 & 10 & 15 & 20 & 25 & 1 & 5 & 10 & 15 & 20 & 25 & 1 & 5 & 10 & 15 & 20 & 25 \\
\hline
KISSME\footnote{The code is downloaded from \url{http://lrs.icg.tugraz.at/research/kissme/}.} & 18.0 & 45.3 & 59.7 & 68.3 & 74.5 & 79.5 & 20.4 & 48.7 & 61.9 & 70.9 & 76.5 & 81.3 & 15.8 & {\bf 45.0} & 58.5 & {\bf 68.8} & 73.8 & 77.5 \\
KISSME + SM   & {\bf 21.5} & {\bf 48.7} & {\bf 66.1} & {\bf 74.3} & {\bf 80.0} & {\bf 84.0} & {\bf 22.9} & {\bf 50.0} & {\bf 63.3} & {\bf 71.1} & {\bf 76.6} & {\bf 81.6} & {\bf 16.8} & 42.3 & {\bf 60.5} & 68.3 & {\bf 76.3} & {\bf 81.5} \\
\hline
\textbf{Ours:} {\sc PRISM-\rmnum{1}} & {\bf 35.9} & {\bf 66.5} & {\bf 81.0} & {\bf 87.2} & {\bf 89.3} & 90.4 & {\bf 36.7} & {\bf 66.6} & {\bf 80.6} & {\bf 86.1} & 87.9 & 89.9 & {\bf 34.5} & 60.0 & 71.8 & 76.0 & 83.3 & 87.0 \\
\textbf{Ours:} {\sc PRISM-\rmnum{2}} & 32.9 & 64.3 & 77.2 & 83.7 & 88.2 & {\bf 90.5} & 32.2 & 62.2 & 74.7 & 83.7 & {\bf 88.0} & {\bf 91.7} & 29.5 & {\bf 60.8} & {\bf 76.3} & {\bf 81.8} & {\bf 87.8} & {\bf 90.3} \\
\textbf{Ours:} {\sc PRISM-\rmnum{3}} & 34.0 & 63.6 & 75.6 & 83.4 & 87.1 & 89.7 & 33.2 & 62.0 & 74.2 & 81.4 & 84.8 & 90.5 & 31.8 & 59.3 & 74.8 & 80.3 & 84.3 & 88.0 \\ 
\hline
\end{tabular}
\end{table*}
\end{savenotes}

For multi-shot learning (see the definition in Section \ref{ssec:rw}), since VIPeR does not have multiple images per person, we compare our method with others on CUHK01 and iLIDS-VID only, and list the comparison results in Table \ref{tab:ms}. Clearly, PRISM beats the state-of-the-art significantly by {\bf 36.7\%} on CUHK01, and {\bf 27.5\%} on iLIDS-VID, respectively, at rank-1. Note that even compared with the best fusion method \cite{2015arXiv150301543P} on CUHK01, our method outperforms it by {\bf 14.7\%} at rank-1. Our multi-shot CMC curves on CUHK01 are also shown in Fig. \ref{fig:single-shot}(b) for comparison. 

The improvement of our method for multi-shot learning mainly comes from the multi-instance setting of our latent spatial kernel in Eq. \ref{eqn:k}. It has been clearly demonstrated as we compare our performances using single-shot learning and multi-shot learning on CUHK01. By averaging over all the gallery images for one entity in multi-shot learning, the visual word co-occurrence model constructed is more robust and discriminative than that for single-shot learning, leading to significant performance improvement.

\subsection{Robust Person Re-identification}

In this experiment, we would like to demonstrate the robustness of our method by including the missing match scenarios for re-id. Here we compare different methods only on VIPeR for the demonstration purpose.

We utilize KISSME \cite{lrs:icg:koestinger12a} to do the comparison, which includes 5 different metric learning methods, namely, KISSME \cite{lrs:icg:koestinger12a}, MAHAL (\ie Mahlanobis distance learning), ITML \cite{davis-et-al-icml-2007}, LDML \cite{GVS09}, and LMNN \cite{Weinberger:2009:DML:1577069.1577078}. These metric learning methods learn the similarities between image pairs, which are equivalent to $\mathbf{w}^T\phi(\mathbf{x}_{ij})$ in Eq. \ref{eqn:test}. Then we apply our structured matching (SM for short) in Eq. \ref{eqn:test} on top of each method above by utilizing these image pair similarities for comparison. 

We first simulate the \reid scenario where every probe has its match in the gallery set, but not all the galleries have matches in the probe set. Fig. \ref{fig:open-world} shows our comparison results using (a) 158 probes, (b) 79 probes, and (c) 40 probes, respectively, with 316 entities in the gallery set. As we see, for all the 5 metric learning methods, structured matching helps improve their performances, in general, under different settings. PRISM always performs best among all the methods.

Table \ref{tab:open-world} summarizes the numbers at different ranks for KISSME, KISSME+SM, and our PRISM in Fig. \ref{fig:open-world}, since KISSME and KISSME+SM are the most comparable methods in Fig. \ref{fig:open-world}. At rank-1, PRISM outperforms them significantly by at least {\bf 14.4\%}. As the number of probes decreases, in general, at every rank the matching rates of all the methods degrades. However, as we see, for PRISM the matching rates are much more stable. By comparing these results with those in Table \ref{tab:ss}, we can see that these results are similar, again demonstrating the robustness of our structured matching method.

\begin{figure}[t]
 \begin{center}
 \centerline{\includegraphics[width=0.95\columnwidth]{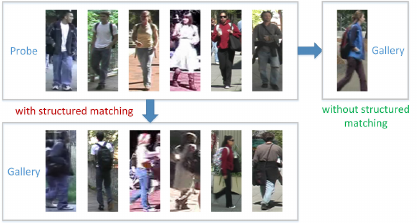}}
% \centerline{\footnotesize{(a)}}
 \end{center}
\vspace{-5mm}
\caption{\footnotesize{Examples of matching result comparison on VIPeR at rank-1 using PRISM with/without structured matching for robust re-id.} The sizes of the probe and gallery sets are 40 and 316, respectively.}\label{fig:open-world-ex}
%\vspace{-3mm}
\end{figure}

%\begin{savenotes}
%\begin{table}[t]\scriptsize\centering
%\begin{tabular}{|l||lll|}
%\hline
%& 158 probes & 79 probes & 40 probes \\
%\hline
%{\sc PRISM-\rmnum{1}} & 158.9 & 35.7 & 1.3 \\
%{\sc PRISM-\rmnum{2}} & 113.7 & 33.8 & 1.4 \\
%{\sc PRISM-\rmnum{3}} & 16.2 & 26.0 & 1.3  \\ 
%\hline
%\end{tabular}
%\caption{\footnotesize{Average storage and computational time for our PRISM.}}\label{tab:time}\vspace{-3mm}
%\end{table}
%\end{savenotes}

\begin{savenotes}
\begin{table}[t]\centering
\caption{\footnotesize{Average matching accuracy (\%) for robust \reid on VIPeR.}}\label{tab:open-world2}
\setlength\tabcolsep{3pt}
\begin{tabular}{|l|ll|lll|}
\hline
\# probe & KISSME & KISSME+SM & {\sc PRISM-\rmnum{1}} & {\sc PRISM-\rmnum{2}} & {\sc PRISM-\rmnum{3}} \\
\hline
158 & 14.9 & {\bf 20.6} & 20.9 & 20.4 & {\bf 21.0} \\
79 & 15.0 & {\bf 16.5} & {\bf 20.8} & 20.2 & {\bf 20.8} \\
40 & 14.4 & {\bf 15.1} & 22.5 & 22.2 & {\bf 22.6} \\
\hline
\end{tabular}
\end{table}
\end{savenotes}

%\subsection{Cross-Domain Learning}
We display representative matching results at rank-1 in Fig. \ref{fig:open-world-ex} using PRISM with/without structured matching for robust re-id. As we see, without structured matching all the probes are matched with the same entity in the gallery, inducing incorrect matches. However, structured matching can correct this type of errors, find their true matches, and thus improve the matching rates. 

Next we simulate another \reid scenario where not all the probes/galleries have matches in the gallery/probe sets. This is a common situation in \reid where missing matches occur all the time. During testing, we randomly select 158/79/40 probes to be matched with randomly selected 158 galleries, and list in Table \ref{tab:open-world2} the results in terms of average {\em matching accuracy}, \ie, in the probe set the total number of true positives (true matches) and true negatives (true non-matches) divided by the total number of entities. Still structured matching helps improve the performance, and PRISM achieves the best.

%In summary, the integration of both structured matching and structured learning of similarity models in PRISM is very suitable for open-world {\it re-id}.

\subsection{Storage \& Computational Time}\label{ssec:ct}

Storage ($S_t$ for short) and computational time during testing are two critical issues in real-world applications. In our method, we only need to store the image descriptors for calculating similarities between different entities. The computational time can be divided into three parts: (1) image descriptors $T_1$, (2) entity-matching similarities $T_2$, and (3) entity-level structured matching $T_3$. We do not consider the time for generating Color+SIFT features, since we directly use the existing code without any control.

We record the storage and computational time using 500 visual words for both probe and gallery sets on VIPeR. The rest of the parameters are the same as described in Section \ref{sec:implementation}. Roughly speaking, the storage per data sample and computational time are linearly proportional to the size of images and number of visual words. Our implementation is based on unoptimized MATLAB code. Numbers are listed in Table \ref{tab:time} for identifying the matches between 316 probes and 316 galleries, including the time for saving and loading features. Our experiments were all run on a multi-thread CPU (Xeon E5-2696 v2) with a GPU (GTX TITAN). The method ran efficiently with very low demand for storage.

\begin{savenotes}
\begin{table}[t]\centering
\caption{\footnotesize{Average storage and computational time for our PRISM.}}\label{tab:time}
\setlength\tabcolsep{3pt}
\begin{tabular}{|l||llll|}
\hline
& $S_t$ (Kb) & $T_1$ (ms) & $T_2$ (ms) & $T_3$ (s) \\
\hline
{\sc PRISM-\rmnum{1}} & 158.9 & 77.4 & 1.3 & 1.3 \\
{\sc PRISM-\rmnum{2}} & 113.7 & 73.3 & 1.4 & 1.5 \\
{\sc PRISM-\rmnum{3}} & 16.2 & 56.4 & 1.3 & 1.3 \\ 
\hline
\end{tabular}
\end{table}
\end{savenotes}

\section{Conclusion}\label{sec:con}

In this paper, we propose a structured matching based method for \reid in the contexts of (1) single-shot learning, and (2) multi-shot learning. We formulate the core of the \reid problem, \ie entity matching, as a weighted bipartite graph matching problem, and try to predict such graph structures. To handle the huge appearance variation (\eg visual ambiguity and spatial distortion) as well as achieving computational efficiency, we propose a new basis function to capture the visual word co-occurrence statistics. Our experiments on several benchmark datasets strongly demonstrate the power of our PRISM for \reid in both scenarios. Low demand of storage and good computational efficiency indicate that our method can be potentially applied to real-world applications.

Several questions will be considered as our future work. It would be useful to further reduce the computational complexity of calculating our pair-wise latent spatial kernels. One possibility is to modify the learning algorithm by decomposing the weight matrix $\mathbf{w}$ into two separable matrices, because our appearance model can be decomposed into two parts, one from the probe image and the other from the gallery image. Such decomposition will accelerate the computation. Second, it would be interesting to learn the optimal spatial kernels and see how they affect the behavior of our visual word co-occurrence model. Third, it would be also interesting to extend our current structured matching framework to multi-camera settings by adding more constraints on the matched/dismatched entity pairs to enforce the structural information (\eg temporal) in the network.
%Building a {\it re-id} system for natural images using object proposal algorithms (\eg \cite{zhang2011proposal,objectnessBING}) and our model with different classifiers (\eg \cite{zhang2010adamkl,zhang2012efficient,zhang2011learning}) would be interesting as well.

\ifCLASSOPTIONcompsoc
  \section*{Acknowledgments}
  This material is based upon work supported by the U.S. Department of Homeland Security, Science and Technology Directorate, Office of University Programs, under Grant Award 2013-ST-061-ED0001. The views and conclusions contained in this document are those of the authors and should not be interpreted as necessarily representing the official policies, either expressed or implied, of the U.S. Department of Homeland Security.
\else
  \section*{Acknowledgment}

\fi

\ifCLASSOPTIONcaptionsoff
  \newpage
\fi

%{\it i.e.}EEtriggeratref{8}
%{\it i.e.}EEtriggercmd{\enlargethispage{-5in}}


\bibliographystyle{IEEEtran}
\bibliography{egbib}
\end{document}